\documentclass{bmvc2k}

\usepackage{times}
\usepackage{epsfig}
\usepackage{graphicx, multirow, graphbox}
\graphicspath{ {figs/} }
\usepackage{amsmath}
\usepackage{amssymb}
\usepackage[dvipsnames]{xcolor}
\usepackage{enumitem}
\usepackage{cprotect}
\usepackage{gensymb}
\usepackage{stackengine}
\usepackage{rotating}
\usepackage{tabularx}

\def\titlespace{\hspace{0.4cm}}

\title{\titlespace Image Classification with \\ \titlespace Hierarchical Multigraph Networks}

\addauthor{Boris Knyazev* $^{\mbox{\sffamily \scriptsize 1,}}$}{bknyazev@uoguelph.ca}{2}
\addauthor{Xiao Lin}{xiao.lin@sri.com}{3}
\addauthor{Mohamed R. Amer* }{mohamed@robust.ai}{4}
\addauthor{Graham W. Taylor$^{\mbox{\sffamily \scriptsize 1,2,}}$}{gwtaylor@uoguelph.ca}{5}

\addinstitution{
	University of Guelph \\ Canada
}
\addinstitution{
	Vector Institute for Artificial Intelligence \\ Canada \\
}
\addinstitution{
	SRI International \\
	Princeton, NJ, USA
}
\addinstitution{
	Robust.AI \\
}
\addinstitution{
	Canada CIFAR AI Chair
}

\runninghead{B. Knyazev, X. Lin, M.R. Amer, G.W. Taylor}{\small Image Class., Hier. Multigraphs}

\def\eg{\emph{e.g}\bmvaOneDot}

\newcommand\Tstrut{\rule{0pt}{2.6ex}}         % = `top' strut
\newcommand\Bstrut{\rule[-0.9ex]{0pt}{0pt}}   % = `bottom' strut
\newcommand{\std}[1]{{\tiny{$\pm$#1}}}

\newcommand{\densepar}[1]{\textbf{#1}}

\newcommand\blfootnote[1]{%
	\begingroup
	\renewcommand\thefootnote{}\footnote{#1}%
	\addtocounter{footnote}{-1}%
	\endgroup
}

\begin{document}

%%%%%%%%% TITLE

\maketitle
%\thispagestyle{empty}
%%%%%%%%% ABSTRACT
\begin{abstract}
Graph Convolutional Networks (GCNs) are a class of general models that can learn from graph structured data.
Despite being general, GCNs are admittedly inferior to convolutional neural networks (CNNs) when applied to vision tasks, mainly due to the lack of domain knowledge that is hardcoded into CNNs, such as spatially oriented translation invariant filters. However, a great advantage of GCNs is the ability to work on irregular inputs, such as superpixels of images. This could significantly reduce the computational cost of image reasoning tasks. Another key advantage inherent to GCNs is the natural ability to model multirelational data. Building upon these two promising properties, in this work, we show best practices for designing GCNs for image classification; in some cases even outperforming CNNs on the MNIST, CIFAR-10 and PASCAL image datasets.
\end{abstract}
\blfootnote{*Most of this work was done while the authors were at SRI International.}
\vspace{-15pt}

%%%%%%%%% BODY TEXT
\section{Introduction}\label{sec:intro}
\begin{figure}[t]
	\small
	\begin{center}
		\begin{footnotesize}
			\begin{tabular}{ccc}
				{{\includegraphics[width=0.1\textwidth]{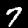}}} & 
				{\includegraphics[width=0.1\textwidth]{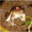}} & {\includegraphics[width=0.15\textwidth]{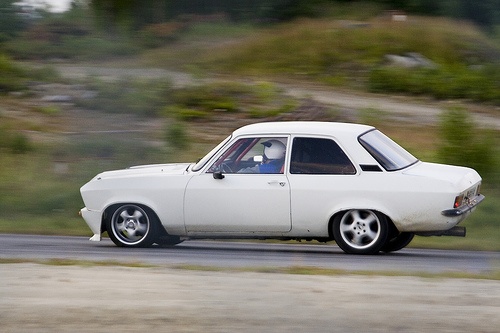}} \\
				\textit{(a)} $28 \times 28$ px & \textit{(b)} $32 \times 32$ px & \textit{(c)} $333 \times 500$ px \\
				{\includegraphics[width=0.1\textwidth]{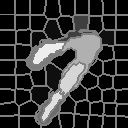}} & {\includegraphics[width=0.1\textwidth]{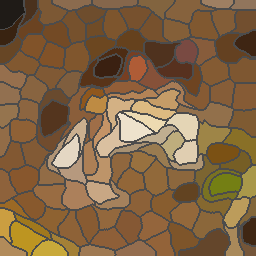}} & {\includegraphics[width=0.15\textwidth]{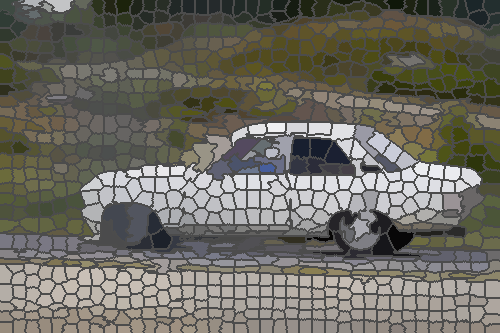}} \\
				\textit{(d)} $N=75$ & \textit{(e)} $N=150$ & \textit{(f)} $N=1000$ \\
			\end{tabular}
		\end{footnotesize}
	\end{center}
	\caption{\small Examples of the original images \textit{(a-c)}, defined on a regular grid, and their superpixel representations \textit{(d-f)} for MNIST \textit{(a,d)}, CIFAR-10 \textit{(b,e)} and PASCAL \textit{(c,f)}; $N$ is the number of superpixels (nodes in our graphs). GCNs can learn both from images and superpixels due to their flexibility, whereas standard CNNs can learn only from images defined on a regular grid \textit{(a-c)}.}
	\label{fig:tasks}
	\vspace{-10pt}
\end{figure}

In image recognition, input data fed to models tend to be high dimensional. Even for tiny MNIST~\cite{lecun1998gradient} images, the input is $28\times28 \text{px} =784$ dimensional and for larger PASCAL~\cite{everingham2010pascal} images it explodes to $333\times500 \text{px} \approx 165,000$ dimensions (Figure~\ref{fig:tasks}).
Learning from such a high-dimensional input is challenging and requires a lot of labelled data and regularization. Convolutional Neural Networks (CNNs) successfully address these challenges by exploiting the properties of shift-invariance, locality and compositionality of images~\cite{bronstein2017geometric}. We consider an alternative approach and instead reduce the input dimensionality. One simple way to achieve that is downsampling. But in this case, we may lose vital structural information, so to better preserve it, we extract superpixels~\cite{achanta2012slic, liang2016semantic}. Representing a set of superpixels such that CNNs could digest and learn from them is non-trivial. To that end, we adopt a strategy from chemistry, physics and social networks, where structured data are expressed by graphs~\cite{bronstein2017geometric, hamilton2017representation, battaglia2018relational}. By defining operations on graphs analogous to spectral~\cite{bruna2013spectral} or spatial~\cite{kipf2016semi} convolution, Graph Convolutional Networks (GCNs) extend CNNs to graph-based data, and show successful applications in graph/node classification~\cite{velickovic2017graph, simonovsky2017dynamic, monti2017geometric, fey2018splinecnn} and link prediction~\cite{schlichtkrull2018modeling}.\looseness=-1

The challenge of generalizing convolution to graphs is to have anisotropic filters (such as edge detectors). Anisotropic models, such as MoNet~\cite{monti2017geometric} and SplineCNN~\cite{fey2018splinecnn}, rely on coordinate structure, work well for various vision tasks, but are often too computationally expensive and suboptimal for graph problems, in which the coordinates are not well defined~\cite{knyazev2018spectral}. While these and other general models exist~\cite{gilmer2017neural, battaglia2018relational}, we rely on widely used graph convolutional networks (GCNs)~\cite{kipf2016semi} and their multiscale extension, Chebyshev GCNs (ChebyNets)~\cite{defferrard2016convolutional} that enjoy an explicit control of receptive field size.

Our focus is on multigraphs, those graphs that are permitted to have multiple edges, \eg two objects can be connected by edges of different types~\cite{lu2016visual}: (\verb+is_part_of+, \verb+action+, \verb+distance+, etc.). Multigraphs enable us to model spatial and hierarchical structure inherent to images. By exploiting multiple relationships, we can capture global patterns in input graphs, which is hard to achieve with a single relation, because most GCNs aggregate information in a small local neighbourhood and simply increasing its size, as in ChebyNet, can quickly make the representation too entangled (due to averaging over too many features). Hence, methods such as Deep Graph Infomax~\cite{velivckovic2018deep} were proposed. Using multigraph networks is another approach to increase the receptive field size and disentangle the representation in a principled way, which we show to be promising.

In this work, we model images as sets of superpixels (Figure~\ref{fig:tasks}) and formulate image classification as a multigraph classification problem (Section~\ref{sec:model}).
We first overview graph convolution (Section~\ref{sec:multigraph}) and then adopt and extend relation type fusion methods from~\cite{knyazev2018spectral} to improve the expressive power of our GCNs (Section~\ref{sec:edge_fusion_methods}). To improve classification accuracy, we represent an image as a multigraph and propose learnable (Section~\ref{sec:learn_edges}) and hierarchical (Section~\ref{sec:hierarchy_images}) relation types that we fuse to obtain a final rich representation of an image. On a number of experiments on the MNIST, CIFAR-10, and PASCAL image datasets, we evaluate our model and show a significant increase in accuracy, outperforming CNNs in some tasks (Section~\ref{sec:experiments}).

%
%--------------------------------------------------------------------
%---------------------------Multigraph Convolution-------------------
%--------------------------------------------------------------------
%
\section{Graph Convolution}\label{sec:multigraph}	

We consider undirected graphs ${\cal G =(V}, A)$ with $N$ nodes $\cal V$ and edges with values in the range $[0, 1]$ represented as an adjacency matrix $A \in \mathbb{R}^{N \times N}$. Nodes $v_i \in \cal V$ usually represent specific semantic concepts such as objects in images~\cite{prabhu2015attribute}. Nodes can also denote abstract blocks of information with common properties, such as superpixels in our case. Edges $A_{ij}$ ($i,j \in [1, N]$) define the relationships and scope of which node effects propagate.

Convolution is an essential computational block in graph networks, since it permits the gradual aggregation of information from neighbouring nodes.
Following~\cite{kipf2016semi}, for some $C$-dimensional features over nodes $X \in \mathbb{R}^{N \times C}$ and trainable filters $\Theta \in \mathbb{R}^{C \times F}$, convolution on a graph $\cal G$ can be defined as:
\begin{equation}
\label{eq:graph_conv}
Y = \bar{X} \Theta,
\end{equation}
\noindent where $\bar{X} = \tilde{D}^{-1/2} \tilde{A} \tilde{D}^{-1/2} X$ are features averaged over one-hop ($K=1$) neighbourhoods, $\tilde{A} = A + I$ is an adjacency matrix with self-loops, $\tilde{D}_{ii} = \sum_j \tilde{A}_{ij}$ is a diagonal matrix with node degrees, and $I$ is an identity matrix. We employ this convolution to build graph convolutional networks (GCNs) in our experiments.
This formulation is a particular case of a more general approximate spectral graph convolution~\cite{defferrard2016convolutional}, in which case $\bar{X} \in \mathbb{R}^{N \times CK}$ are multiscale features averaged over $K$-hop neighbourhoods. Multiscale (multihop) graph convolution is used in our experiments with ChebyNet.

\begin{figure}[]
 	\begin{center}
		{\includegraphics[width=0.75\textwidth,trim={1.5cm 13cm 8cm 2cm}, clip]{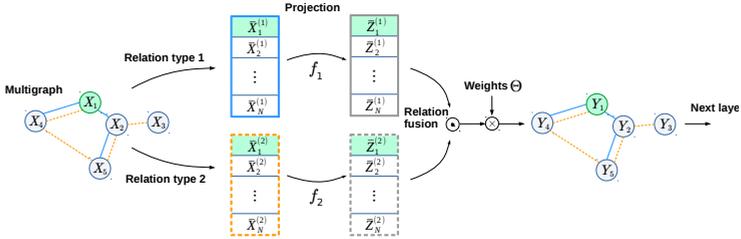}}
 	\end{center}
 	\vspace{-10pt}
 	\caption{\small
 		An example of the ``PC'' relation type fusion based on a trainable projection (Eq.~\ref{eq:proj_concat_conv}). We first project features onto a common multirelational space, where we fuse them using a fusion operator, such as summation or concatenation.
		In this work, relation types 1 and 2 can denote spatial and hierarchical (or learned) edges. We also allow for three or more relation types.} 	
 	\label{fig:edge_fusion}
\end{figure}

%
%--------------------------------------------------------------------
%-------------------Extending Graphs and Edge Relations--------------
%--------------------------------------------------------------------
%
%
\section{Multigraph Convolution}
\label{sec:edge_fusion_methods}
In graph convolution (Eq.~\ref{eq:graph_conv}), the graph adjacency matrix $A$ encodes a single ($R=1$) relation type between nodes. We extend Eq.~\ref{eq:graph_conv} to a multigraph: a graph with multiple ($R \geq 1$) edges (relations) between the same nodes, represented as a set of adjacency matrices $\{A^{(r)}\}_1^R$.
Previous work used concatenation- or decomposition-based schemes~\cite{schlichtkrull2018modeling, chen2018iterative} to fuse multiple relations. Instead, to capture more complex multirelational features, we adopt and extend recently proposed fusion methods from~\cite{knyazev2018spectral}.

Two of the methods from~\cite{knyazev2018spectral} have certain limitations preventing us to adopt them directly in this work. In particular, the approach based on multidimensional Chebyshev polynomials is often infeasible to compute and was not shown to be superior in downstream tasks. In our experience, we also found that the multiplicative fusion was unstable to train.
To that end, motivated by the success of multilayer projections in Graph Isomorphism Networks~\cite{xu2018how}, we propose two simple yet powerful fusion methods (Figure~\ref{fig:edge_fusion}).

Given features $\bar{X}^{(r)}$ corresponding to a relation type $r$, in the first approach we concatenate features for all relation types and then transform them using a two layer fully-connected network $f$ with $F_{\text{hid}}$ hidden units and the ReLU activation:
\begin{equation}
\label{eq:concat_proj_conv}
Y = f_{\text{CP}}([\bar{X}^{(0)}, \bar{X}^{(1)},...,\bar{X}^{(R-1)}]),
\end{equation}
where the first and second layers of $f_{\text{CP}}$ have $CKR \times F_{\text{hid}}$ and $F_{\text{hid}} \times F$ trainable parameters respectively, ignoring a bias.
The second approach, illustrated in Figure~\ref{fig:edge_fusion}, is similar to multiplicative/additive fusion~\cite{knyazev2018spectral}, but instead of multiplication/addition we use concatenation:
\begin{equation}
\label{eq:proj_concat_conv}
Y = [\bar{Z}^{(0)}, \bar{Z}^{(1)}, ... , \bar{Z}^{(R-1)}]  \Theta,
\end{equation}
where $Z^{(r)}=f_r(\bar{X}^{(r)})$ and $f_r$ is a single layer fully-connected network with $F_{\text{hid}}$ output units followed by a nonlinearity, so that $f_r$ and $\Theta$ have $CK \times F_{\text{hid}}$ and $RF_{\text{hid}} \times F$ trainable parameters respectively, ignoring a bias. Hereafter, we denote the first approach as CP (concatenation followed by projection) and the second as PC (projection followed by concatenation).

%
%--------------------------------------------------------------------
%----------------Multigraph Convolutional Networks-------------------
%--------------------------------------------------------------------
%
\section{Multigraph Convolutional Networks}\label{sec:model}

Image classification was recently formulated as a graph classification problem in~\cite{defferrard2016convolutional,monti2017geometric,fey2018splinecnn}, who considered small-scale image classification problems such as MNIST~\cite{lecun1998gradient}. In this work, we present a model that scales to more complex and larger image datasets, such as PASCAL VOC 2012~\cite{everingham2010pascal}.
We follow~\cite{monti2017geometric} and compute SLIC~\cite{achanta2012slic} superpixels for each image and build a graph, in which each node $v_i$ corresponds to a superpixel and edges $A_{ij}$ ($i, j \in [1, N]$) are computed based on the Euclidean distance between the coordinates $p_i \in \mathbb{R}^2$ and $p_j \in \mathbb{R}^2$ of their centres of masses using a Gaussian with some fixed width $\sigma$:
\begin{equation}
\label{eq:spatial_edge}
A^{(r_{\text{spatial}})}_{ij} = e^{(-\frac{||p_i - p_j||^2}{2\sigma^2})}.
\end{equation}
A frequent assumption of current GCNs is that there is at most one edge between any pair of nodes in a graph. This restriction is usually implied by datasets with such structure, so that in many datasets, graphs are annotated with the single most important relation type. Meanwhile, data is often complex and nodes tend to have multiple relationships of different semantic, physical, or abstract meanings.
Therefore, we argue that there could be other relationships captured by relaxing this restriction and allowing for multiple kinds of edges, beyond those hardcoded in the data (\eg spatial in Eq.~\ref{eq:spatial_edge}).

\subsection{Learning Flat vs.~Hierarchical Edges}
Prior work, \eg~\cite{schlichtkrull2018modeling,bordes2013translating},
proposed methods to learn from multiple edges, but similarly to the methods with a single edge type~\cite{kipf2016semi}, they leveraged only predefined edges in the data.
We formulate a more flexible model, which, in addition to learning from an arbitrary number of relations between nodes (see Section~\ref{sec:edge_fusion_methods}), learns abstract edges jointly with a GCN.

\subsubsection{Flat Learnable Edges}\label{sec:learn_edges}
\vspace{-5pt}

We combine ideas from~\cite{henaff2015deep,velickovic2017graph,simonovsky2017dynamic,battaglia2018relational} and propose to learn a new edge $A^{(r_{\text{learn}, l})}_{ij}$ from any node $v_i$ to node $v_j$ with coordinates $p_i$ and $p_j$ using a trainable similarity function:
\begin{equation}
\label{eq:edge_predict_general}
A^{(r_{\text{learn},l})}_{ij}=f_{\text{edge}}^{(l)}\left( p_i, p_j \right), j \in \mathcal{N}^\epsilon_i,
\end{equation}
where $l \in [1, L]$ indexes relation types; $L$ is the number of learned relation types; and $\mathcal{N}^\epsilon_i$ is a spatial neighbourhood of radius $\epsilon$ around node $v_i$. Using relatively small $\mathcal{N}^\epsilon_i$ limits a model's predictive power, but is important for regularization and to meet computational requirements for larger images from which we extract many superpixels, such as in the PASCAL dataset. We also experimented with feeding node features, such as mean pixel intensities, to this function, but it did not give positive outcomes. Instead, we further regularize the model by constraining the input to be the absolute coordinate differences in lieu of raw coordinates and applying the softmax on top of the predictions:
\begin{equation}
\label{eq:edge_predict}
f_{\text{edge}}^{(l)}\left( p_i, p_j \right) = \text{softmax}\left[ f^{(l)}\left( |p_i - p_j | \right) \right] , j \in \mathcal{N}^\epsilon_i,
\end{equation}
where $f$ is a small two layer fully-connected network with $L$ output units and $f^{(l)}$ denotes taking the $l$-th output.
Using the absolute value makes the filters symmetric (Figure~\ref{fig:pred_edges}), but still sensitive to orientation as opposed to the spatial edge defined in Eq.~\ref{eq:spatial_edge}. The softmax is used to encourage sparse (more localized) edges.

Our approach to predict edges can be viewed as a particular case of generative models of graphs, such as~\cite{simonovsky2018graphvae}. However, the objectives of our model and the latter are orthogonal. Generative models typically make probabilistic predictions to induce diversity of generated graphs, and generation of each edge, node and their attributes should respect other (already generated) edges and nodes in the graph, so that the entire graph becomes plausible.
In our work, we aim to scale our model to larger graphs and assume locality of relationships in visual data, therefore we design a 1) simple deterministic function 2) that makes predictions depending only on local information.

\newcommand{\imgwidth}{0.13\textwidth}
\newcommand{\adjwidth}{0.19\textwidth}
\begin{figure}[b!]
	\begin{center}
		\small
		\begin{tabular}{ccccc}
			%\rotatebox[origin=c]{90}{\parbox{1cm}{\centering \tiny SLIC superpixels}} &
			\includegraphics[width=\imgwidth, align=c]{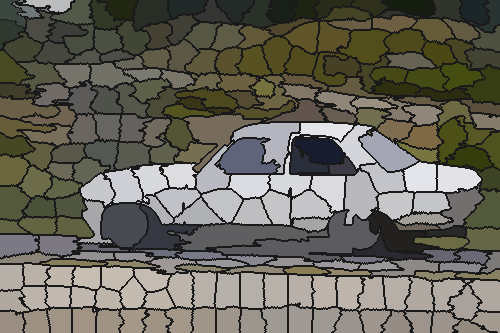} &
			\includegraphics[width=\imgwidth, align=c]{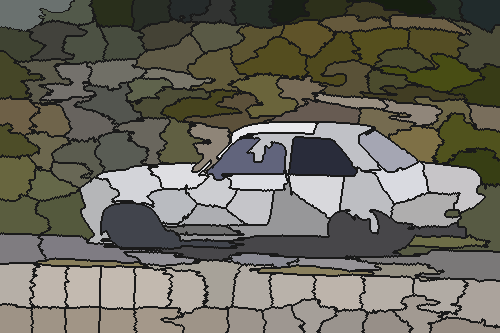} &
			\includegraphics[width=\imgwidth, align=c]{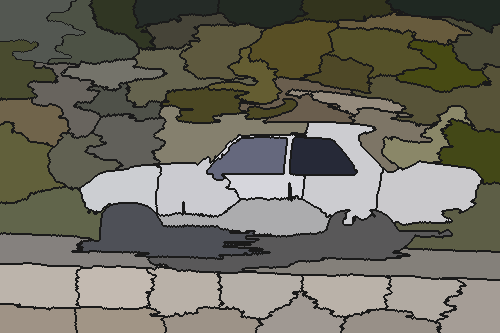} &
			\includegraphics[width=\imgwidth, align=c]{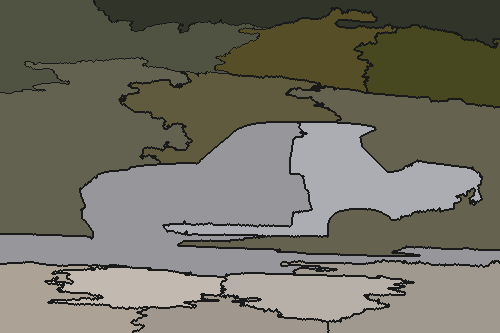} &
			\includegraphics[width=\imgwidth, align=c]{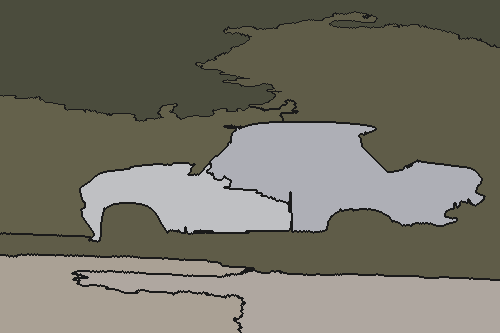} \\
			% (a) 300 sp & (b) 150 sp & (c) 75 sp & (d) 21 sp & (e) 7 sp \\
		\end{tabular}
		\begin{tabular}{cccc}
			\includegraphics[width=\adjwidth, trim={0cm 1cm 0 1cm}, clip]{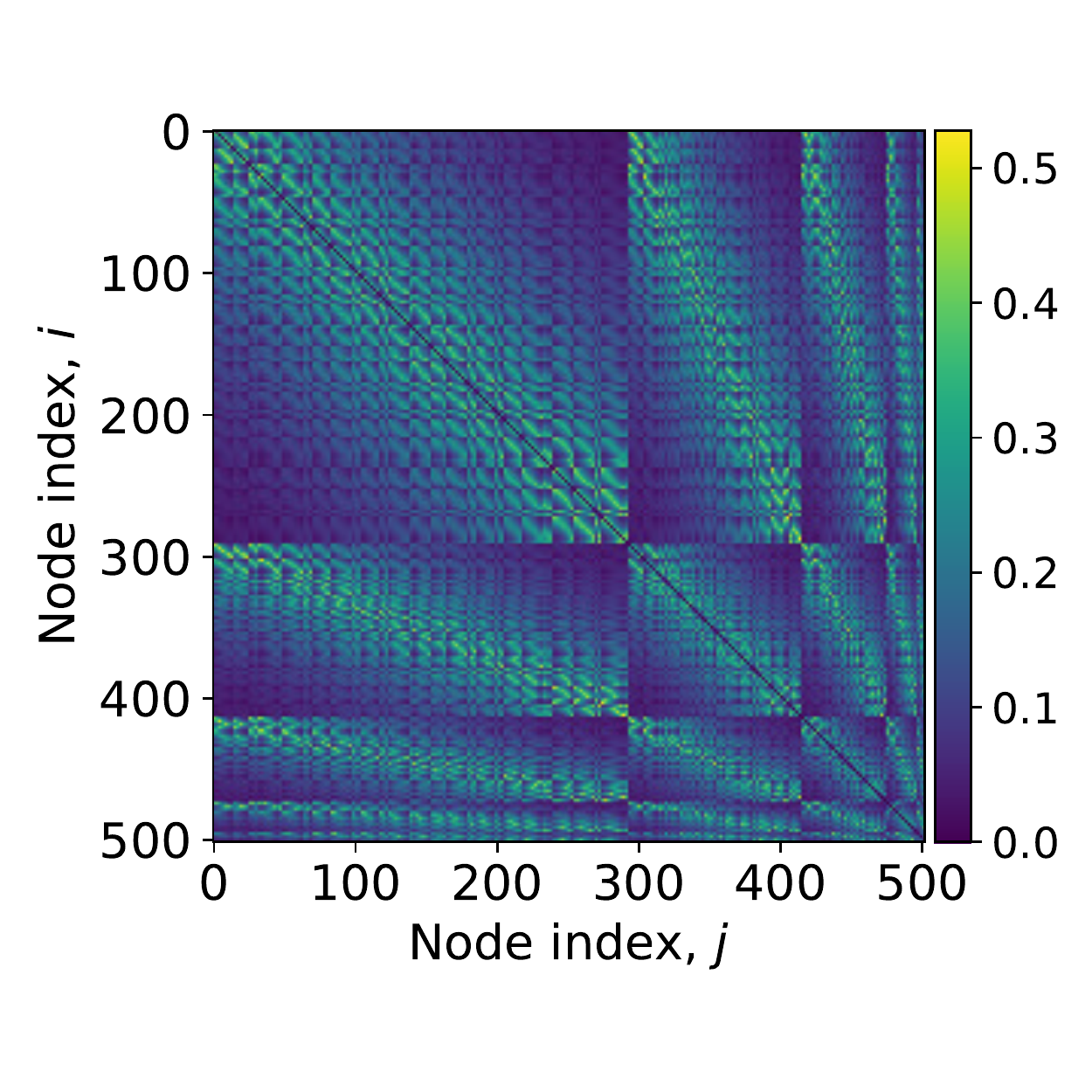} &
			\includegraphics[width=\adjwidth, trim={0cm 1cm 0 1cm}, clip]{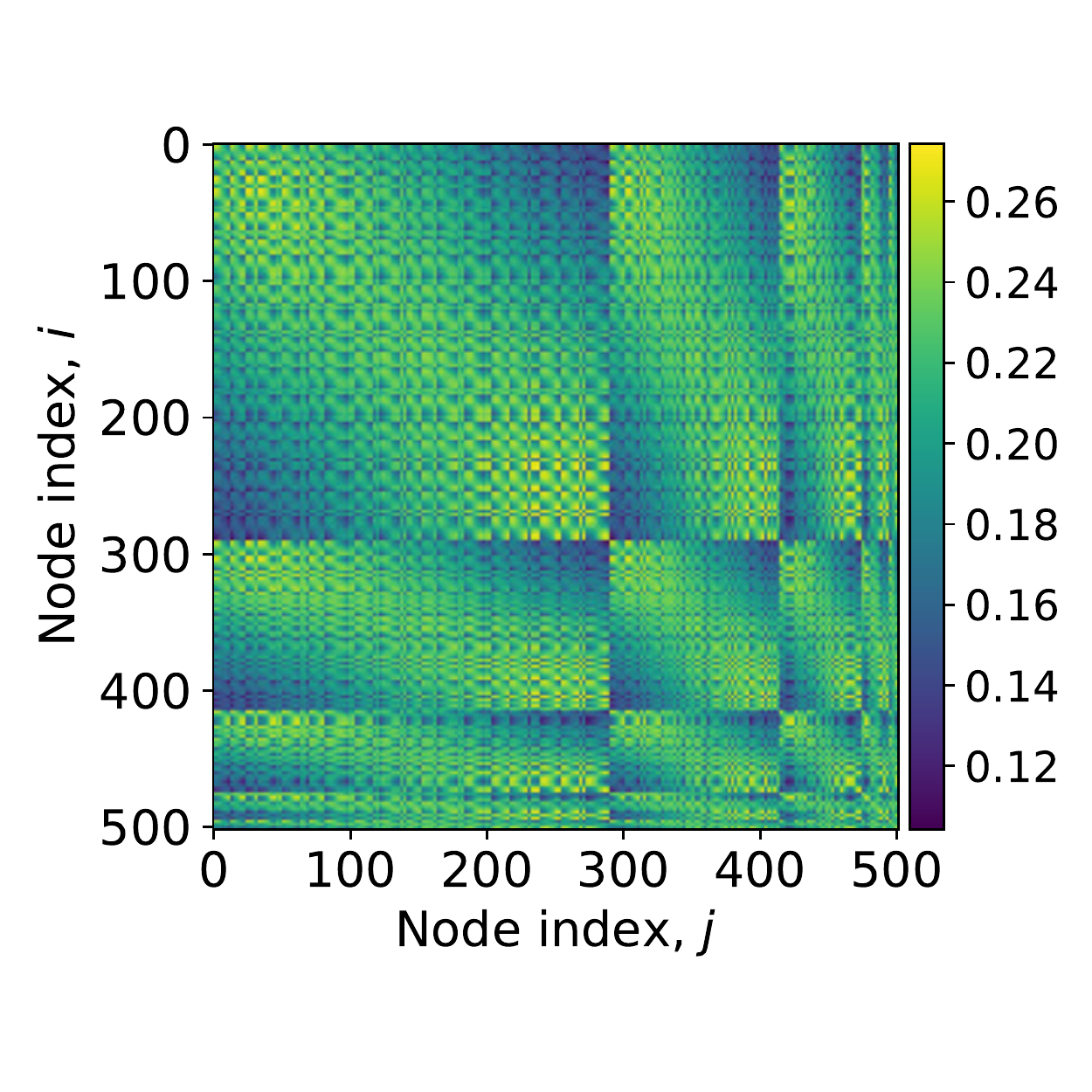} &
			\includegraphics[width=\adjwidth, trim={0cm 1cm 0 1cm}, clip]{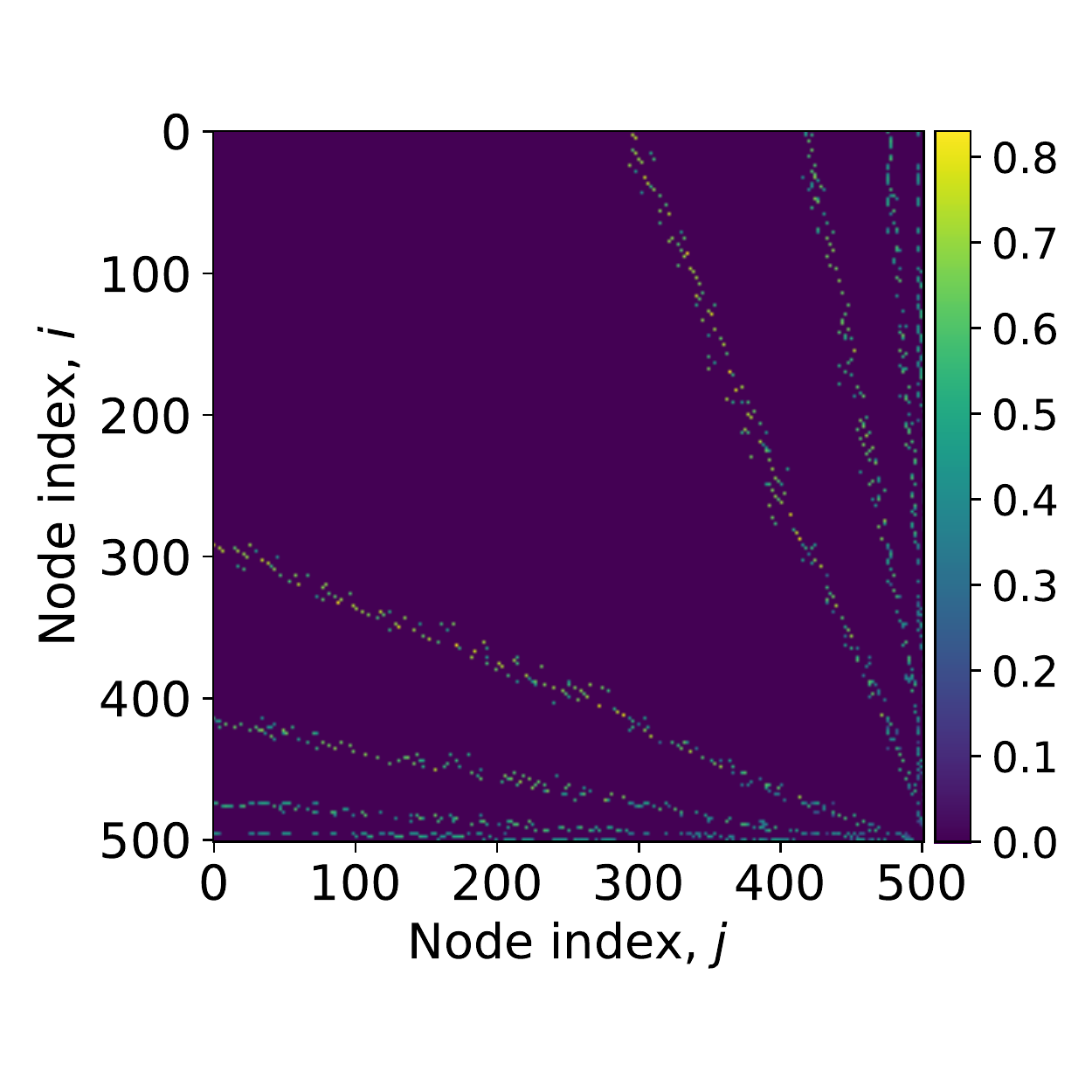} &
			\includegraphics[width=\adjwidth, trim={0cm 1cm 0 1cm}, clip]{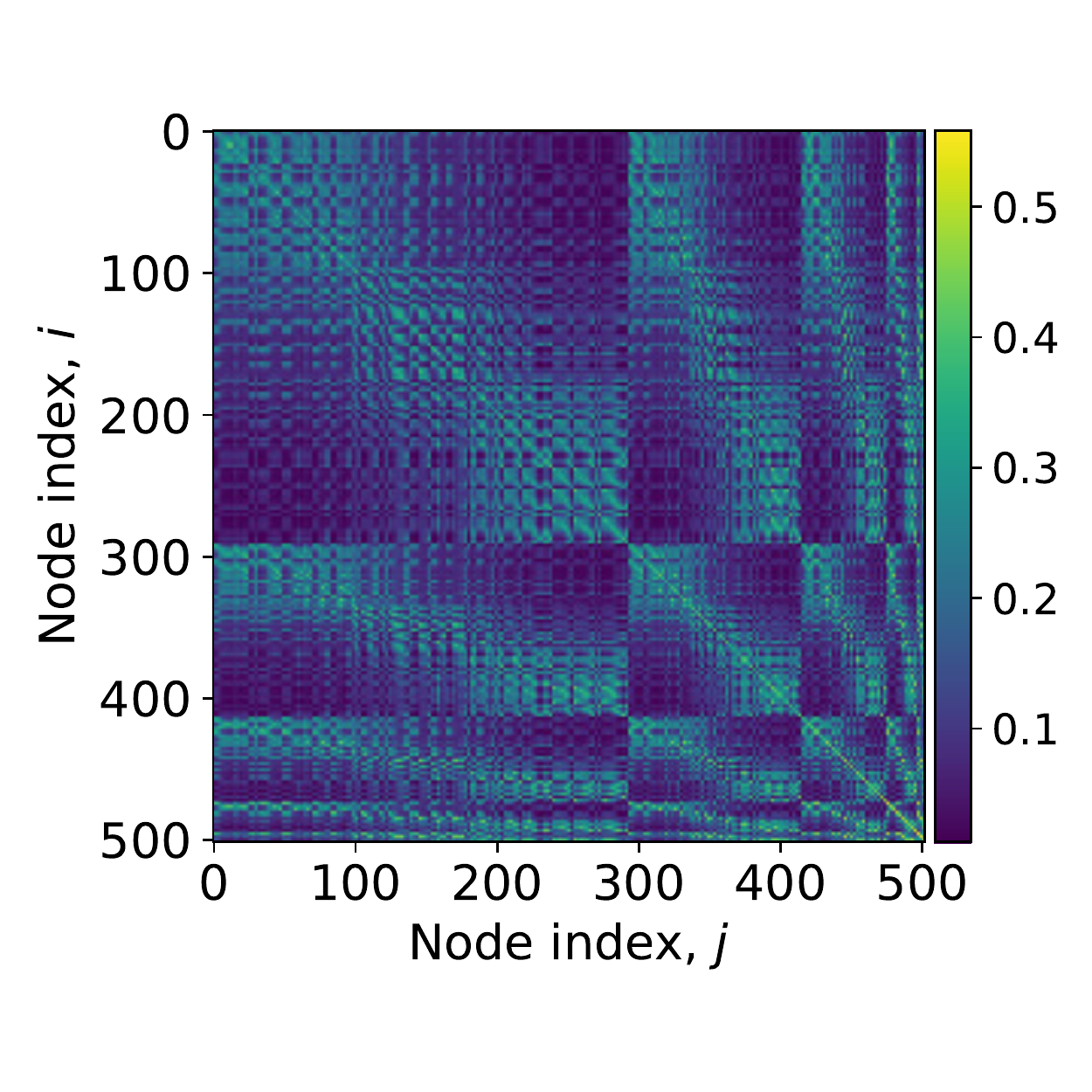} \\
			\textit{(a)} spatial ($K=1$) & \textit{(b)} spatial ($K=7$) & \textit{(c)} hierarchy ($K=1$) & (d) hierarchy ($K=7$)  \\
		\end{tabular}
	\end{center}
	\vspace{-5pt}
	\caption{\small \textit{(top)} We compute superpixels at several scales and combine all of them into a single set. \textit{(bottom}) We then build a graph, where each node corresponds to a superpixel from this set and has features, such as mean RGB color and coordinates of the centres of masses. Using Eq.~\ref{eq:spatial_edge} and~\ref{eq:hier_edge}, we compute spatial \textit{(a)} and hierarchical \textit{(c)} edges. Nodes 0 to 300 correspond to the first level of the hierarchy (first scale of superpixels), and nodes 300 to 400 correspond to the second level, and so forth. Notice that spatial edges \textit{(a)} are created both within and between levels, while hierarchical \textit{(c)} edges exist only between hierarchical levels. \textit{(c, d)} Powers of the adjacency matrices used in a multiscale ChebyNet allow information to diffuse over the graph making it possible to learn filters with more global support.\looseness=-1}
	\label{fig:adjacency_examples}
\end{figure}

\subsubsection{Hierarchical Edges}\label{sec:hierarchy_images}
\vspace{-5pt}
Depending on the number of SLIC superpixels, we can build different levels of image abstraction.
Low levels of abstraction maintain fine-grained details, whereas higher levels mainly preserve global structure. To leverage useful information from multiple levels, we first compute superpixels at several different scales. Then, spatial and learnable relations are computed using Eq.~\ref{eq:spatial_edge} and~\ref{eq:edge_predict}, treating nodes at different scales as a joint set of superpixels. This creates a graph of multiscale image representation (Figure~\ref{fig:adjacency_examples} \textit{(a,b)}).
However, this representation is still flat and, thus, limited. 

To alleviate this, we introduce a novel hierarchical graph model, where child-parent relations are based on intersection over union (IoU) between superpixels $v_i$ and $v_j$ at different scales:
\begin{equation}
\label{eq:hier_edge}
A^{(r_{\text{hier}})}_{ij} = \text{IoU}(v_i, v_j),
\end{equation}
where $A^{(r_{\text{hier}})}$ is an adjacency matrix of hierarchical edges and $A^{(r_{\text{hier}})}_{ij} = 0$ for nodes at the same scale (Figure~\ref{fig:adjacency_examples} \textit{(c,d)}). Using IoU means that child nodes can have multiple parents. To guarantee a single parent, hierarchical superpixel algorithms can be considered~\cite{arbelaez2010contour}.

\subsection{Layer vs.~Global pooling}

\begin{figure}[tbhp]
	\begin{center}
		{\includegraphics[width=0.79\textwidth, trim={1.1cm 15cm 1.8cm 1.7cm}, clip]{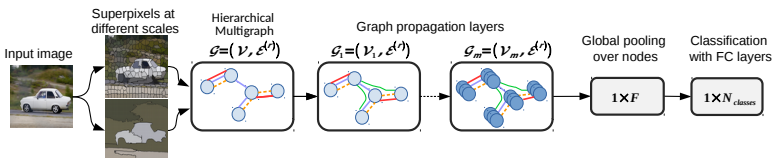}}
	\end{center}
	\vspace{-10pt}
	\caption{\small Image classification pipeline using our model. Each $m\textsuperscript{th}$ graph convolutional layer in our model takes the graph ${\cal G}_m = ({\cal V}_m, {\cal E}^{(r)})$ and returns a graph with the same nodes and edges. Node features become increasingly global after each subsequent layer as the receptive field increases, while edges are propagated without changes. As a result, after several graph convolutional layers, each node in the graph contains information about its immediate neighbours and a large neighbourhood around it. By pooling over nodes we summarize the information collected by each node. Fully-connected layers follow global pooling to perform classification.}
	\label{fig:pipeline}
\end{figure}

Inspired by convolutional networks, previous works~\cite{bruna2013spectral, defferrard2016convolutional,monti2017geometric,simonovsky2017dynamic,fey2018splinecnn,ying2018hierarchical} built an analogy of pooling layers in graphs, for example, using the Graclus clustering algorithm~\cite{dhillon2007weighted}.
In CNNs, pooling is an effective way to reduce memory and computation, particularly for large inputs. It also provides additional robustness to local deformations and leads to faster growth of receptive fields. However, we can build a convolutional network without any pooling layers with similar performance on a downstream task~\cite{springenberg2014striving} --- it just will be relatively slow, since pooling is extremely cheap on regular grids, such as images. In graph classification tasks, the input dimensionality, which corresponds to the number of nodes $N=|\cal V|$, is often very small ($\sim10^2$) and the benefits of pooling are less clear. Graph pooling, such as in~\cite{dhillon2007weighted}, is also computationally intensive since we need to run the clustering algorithm for each graph independently, which limits the scale of problems we can address. Aiming to simplify the model while maintaining classification accuracy, we exclude pooling layers between graph convolutional layers and perform global maximum pooling over nodes following the last conv.~layer. This fixes the size of the penultimate feature vector irrespective of $N$ (Figure~\ref{fig:pipeline}).
%
%--------------------------------------------------------------------
%-----------------------------Experiments----------------------------
%--------------------------------------------------------------------
%
\section{Experiments}\label{sec:experiments}

We evaluate our model on three image-based graph classification datasets: MNIST~\cite{lecun1998gradient}, CIFAR-10~\cite{krizhevsky2009learning} and PASCAL Visual Object Classes 2012 (PASCAL)~\cite{everingham2010pascal}. For each dataset, there is a set of graphs with different numbers of nodes (superpixels), and each graph $\cal G$ has a single categorical label that is to be predicted.
For baselines we use a single ($R=1$) spatial relation type defined using Eq.~\ref{eq:spatial_edge}. For PASCAL we predict multiple categorical labels per image and report mean average precision (mAP).

%\densepar{Datasets.}
MNIST consists of 70k greyscale 28$\times$28px images of handwritten digits.
CIFAR-10 has 60k coloured 32$\times$32px images of 10 categories of animals and vehicles.
PASCAL is a more challenging dataset with realistic high resolution images (typically around $300 \times 500$px) of 20 object categories (people, animals, vehicles, indoor objects). We use standard classification splits, training our model on 5,717 images and reporting results on 5,823 validation images. We note that CIFAR-10 and PASCAL have not been previously considered for graph-based image classification, and in this work we scale our method to these datasets.
In fact, during experimentation we found some other graph convolutional methods unable to scale (see Section \ref{sec:discuss}).

\subsection{Architectural and Experimental Details}\label{sec:arch_exper}

\noindent\textbf{GCNs.}
In all experiments, we train GCNs, ChebyNets and MoNet~\cite{monti2017geometric} with three graph convolutional layers, having 32, 64 and 512 filters with the ReLU activation after each layer followed by global max pooling and a fully-connected classification layer (Figure~\ref{fig:pipeline}). For MNIST, we use a dropout rate~\cite{srivastava2014dropout} of 0.5 before the class.~layer, while for CIFAR-10 and PASCAL instead we employ batch normalization (BN)~\cite{ioffe2015batch} after each convolutional layer.
For edge fusion, projection $f_{\text{CP}}$ in Eq.~\ref{eq:concat_proj_conv} is modelled by a two layer network with $F_{\text{hid}}=64$ hidden units and the ReLU activation between layers. Similarly, projections $f_r(\bar{X}^{(r)})$ in Eq.~\ref{eq:proj_concat_conv} are single layer neural networks with $F_{\text{hid}}=64$ output units followed by the $\tanh$ activation. The edge prediction function in $f_{\text{\text{edge}}}$ (Eq.~\ref{eq:edge_predict}) is a two layer neural network with $32$ hidden units, $L$ output units, and neighbourhood $\mathcal{N}^\epsilon_i$ is set to the $0.2N$ spatially nearest nodes.

\densepar{ConvNets (CNNs).}
To allow fair comparison, we train CNNs with the same number of filters in convolutional layers as GCNs, with filters of size 3 for MNIST, 5 for CIFAR-10 and 7 for PASCAL, max pooling between layers and global average pooling after the last convolutional layer. Deeper and larger CNNs and GCNs can be trained to further improve results. Since images fed to CNNs are defined on a regular grid, adding coordinate features is uninformative, because these features are exactly the same for all examples in the dataset.

\densepar{Low resolution ConvNets.}
GCNs take a relatively low resolution input compared to CNNs: 75 vs.~784 for MNIST, 150 vs.~1024 for CIFAR-10 and 1000 vs.~150,000 for PASCAL. Factors by which it is reduced are 10.5, 6.8 and 150 respectively. Therefore, direct comparison of GCNs to CNNs is unfair. To provide an alternative (yet not perfect) form of comparison, we design experiments with low resolution inputs fed to CNNs. In particular, to match the spatial dimensions of inputs to GCNs and CNNs, for MNIST we reduce the size to $9 \times 9$ px, for CIFAR-10 to $12 \times 12$ px and for PASCAL to $32 \times 32$ px taking into account that the average number of superpixels returned by the SLIC algorithm is often smaller than requested. Admittedly, downsampling using SLIC superpixels is more structural than bilinear downsampling, so GCNs receive a stronger signal, but we believe it is still an interesting experiment. Principal component analysis could be used as a more adequate way to implicitly reduce the input dimensionality for CNNs, but this method is infeasible for large images, such as in PASCAL, while superpixels can be easily computed in such cases. For comparison, we also report results on low resolution images using GCNs.

\densepar{Training.}
We train all models using Adam~\cite{kingma2014adam} with learning rate of $1\mathrm{e}{-3}$, weight decay of $1\mathrm{e}{-4}$, and batch size of 32. For MNIST, the learning rate is decayed after 20 and 25 epochs and models are trained for 30 epochs. For CIFAR-10 and PASCAL, the learning rate is decayed after 35 and 45 epochs and models are trained for 50 epochs.
We train models using four edge fusion methods: concatenation-based baseline, additive fusion~\cite{knyazev2018spectral} and two methods (CP and PC) introduced in Section~\ref{sec:edge_fusion_methods} (see Table~\ref{table:edge_fusions} and Figure~\ref{fig:edge_fusions_sparsity}~\textit{(a)} for a summary).

In each case, we train a model 5 times with different random seeds and report average results and standard deviation. In Table~\ref{table:image_class_results}, we report the best results over different fusion methods superscripting the best fusion method.

\newcommand{\figwidth}{0.10\textwidth}
\begin{figure}[t]
	\centering
	\tiny
	\setlength{\tabcolsep}{1pt}
	\begin{tabular}{|cc|ccccc|ccccc|}
		\hline
		\rotatebox[origin=c]{90}{{Spatial edge}} & 
		{\includegraphics[width=\figwidth, trim={2.5cm 0cm 1cm 0cm}, clip, align=c]{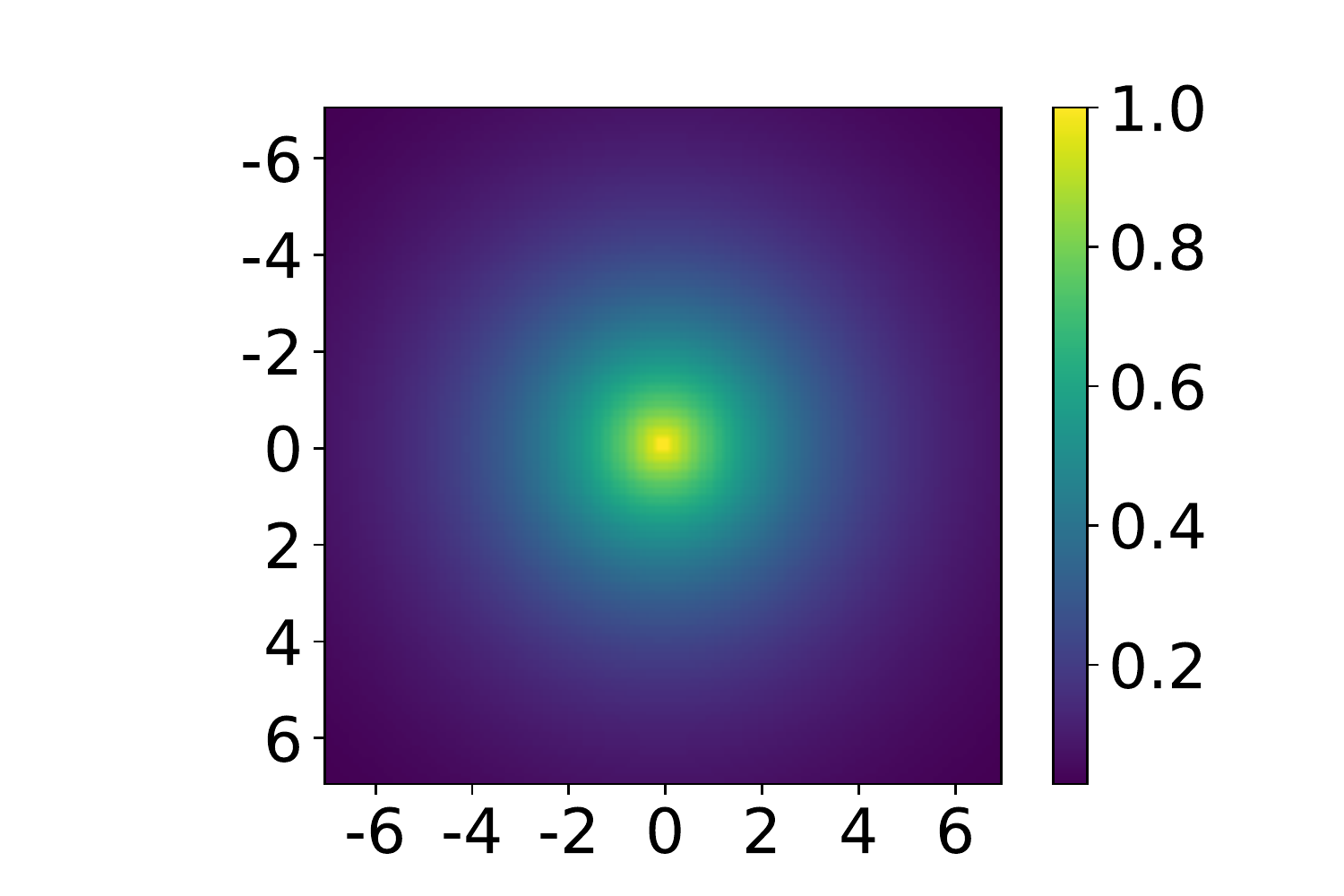}} & 
		\rotatebox[origin=c]{90}{{MNIST}} & 
		{\includegraphics[width=\figwidth, trim={2.5cm 0cm 1cm 0cm}, clip, align=c]{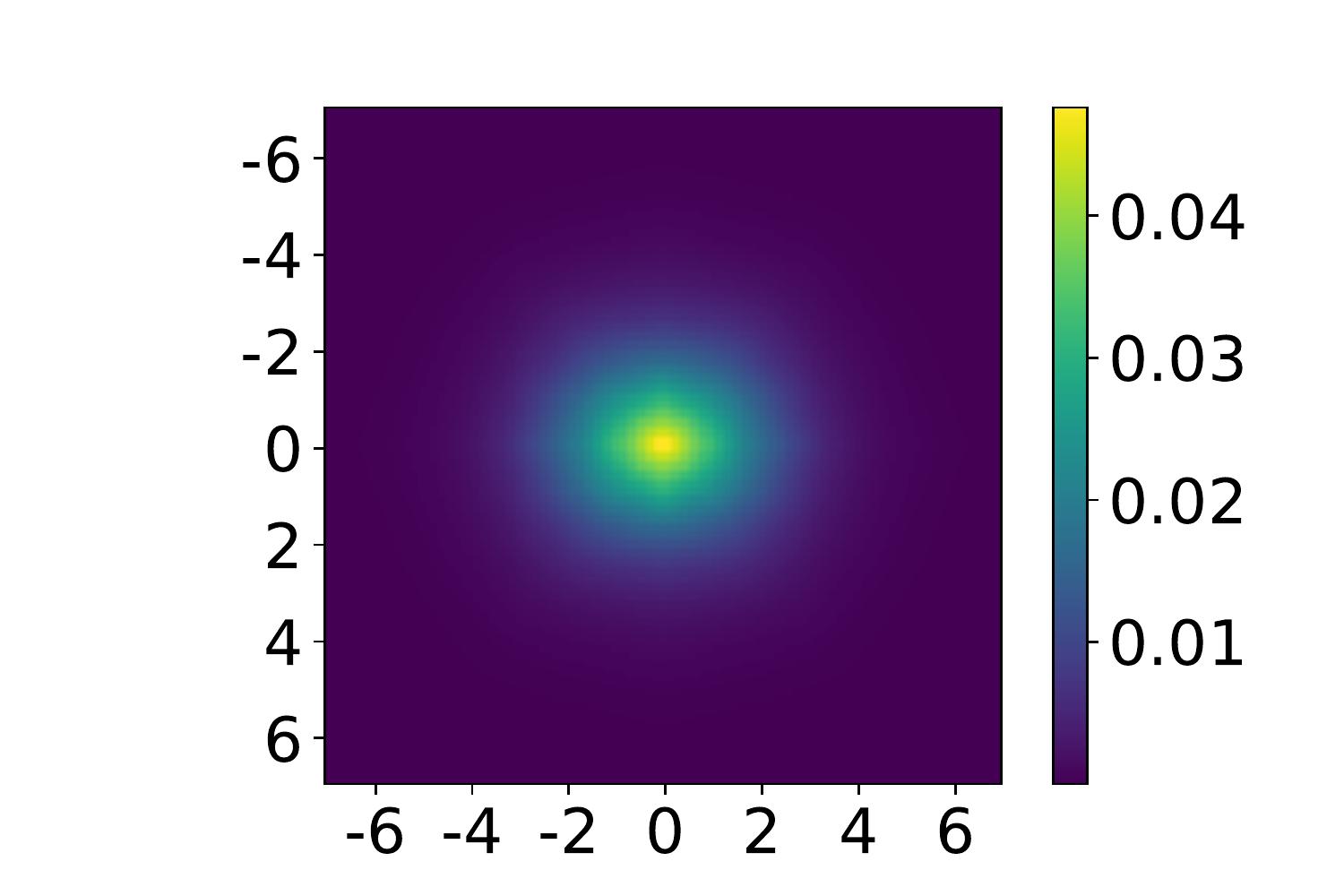}} &
		\includegraphics[width=\figwidth, trim={2.5cm 0cm 1cm 0cm}, clip, align=c]{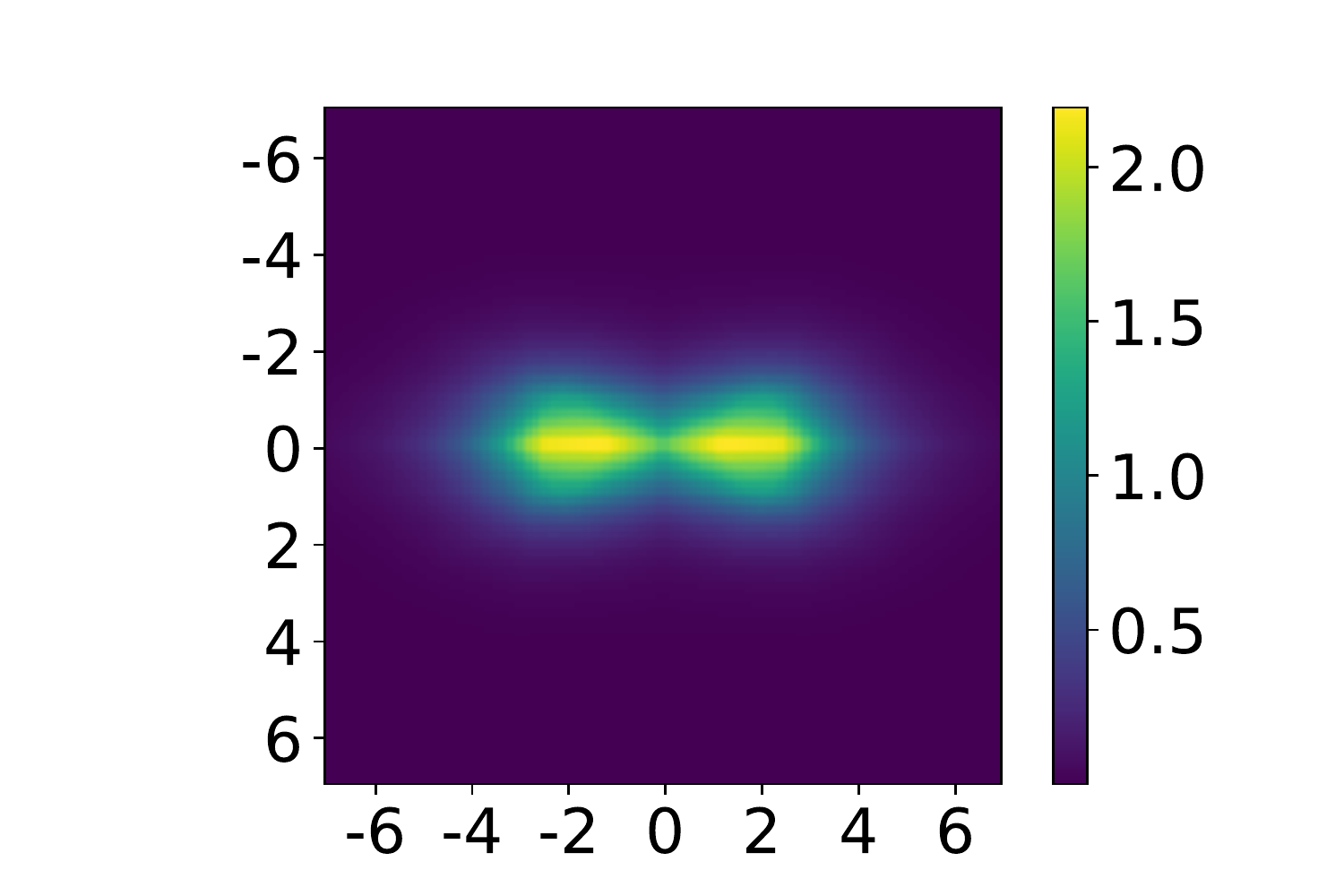} &
		\includegraphics[width=\figwidth, trim={2.5cm 0cm 1cm 0cm}, clip, align=c]{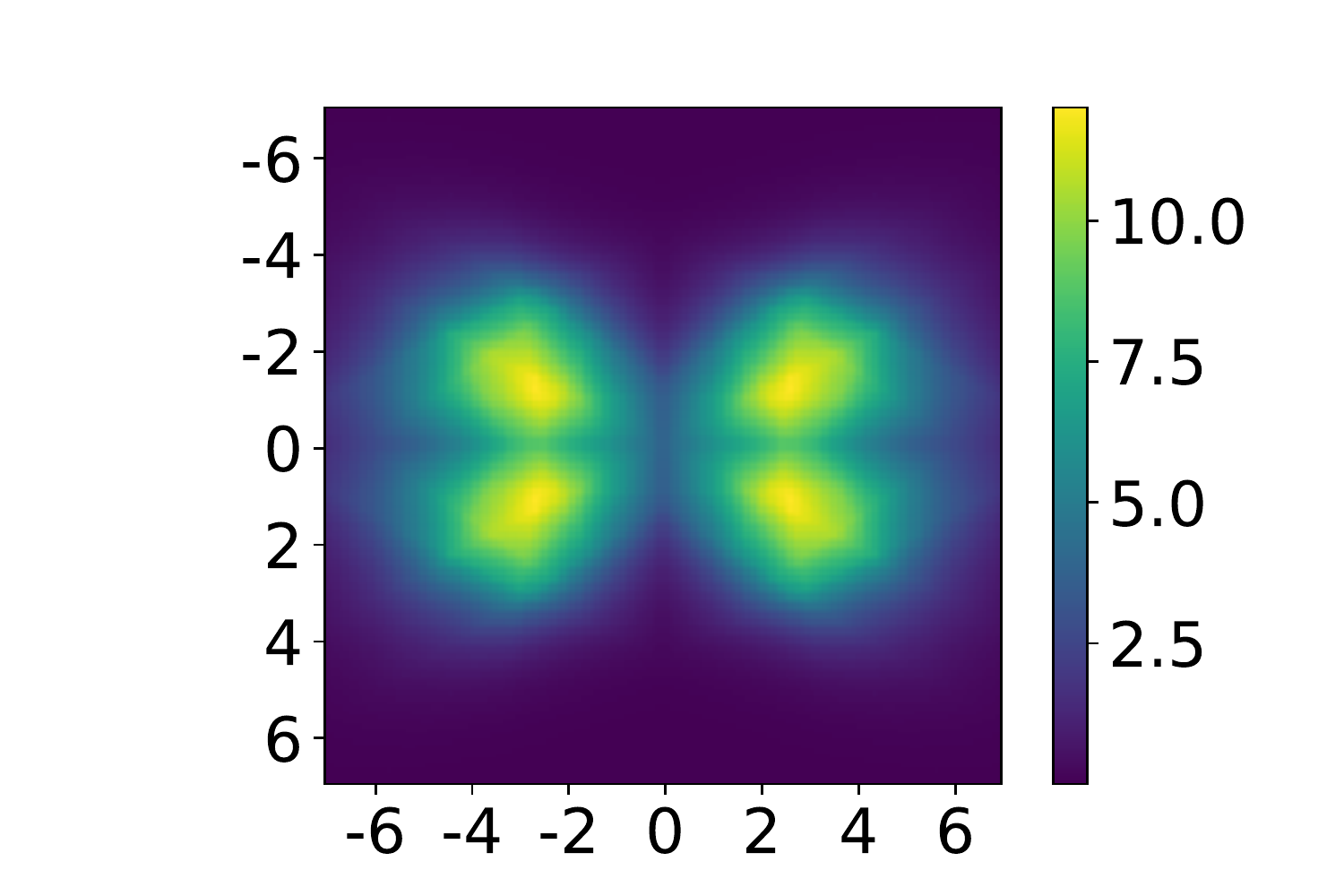} &
		\includegraphics[width=\figwidth, trim={2.5cm 0cm 1cm 0cm}, clip, align=c]{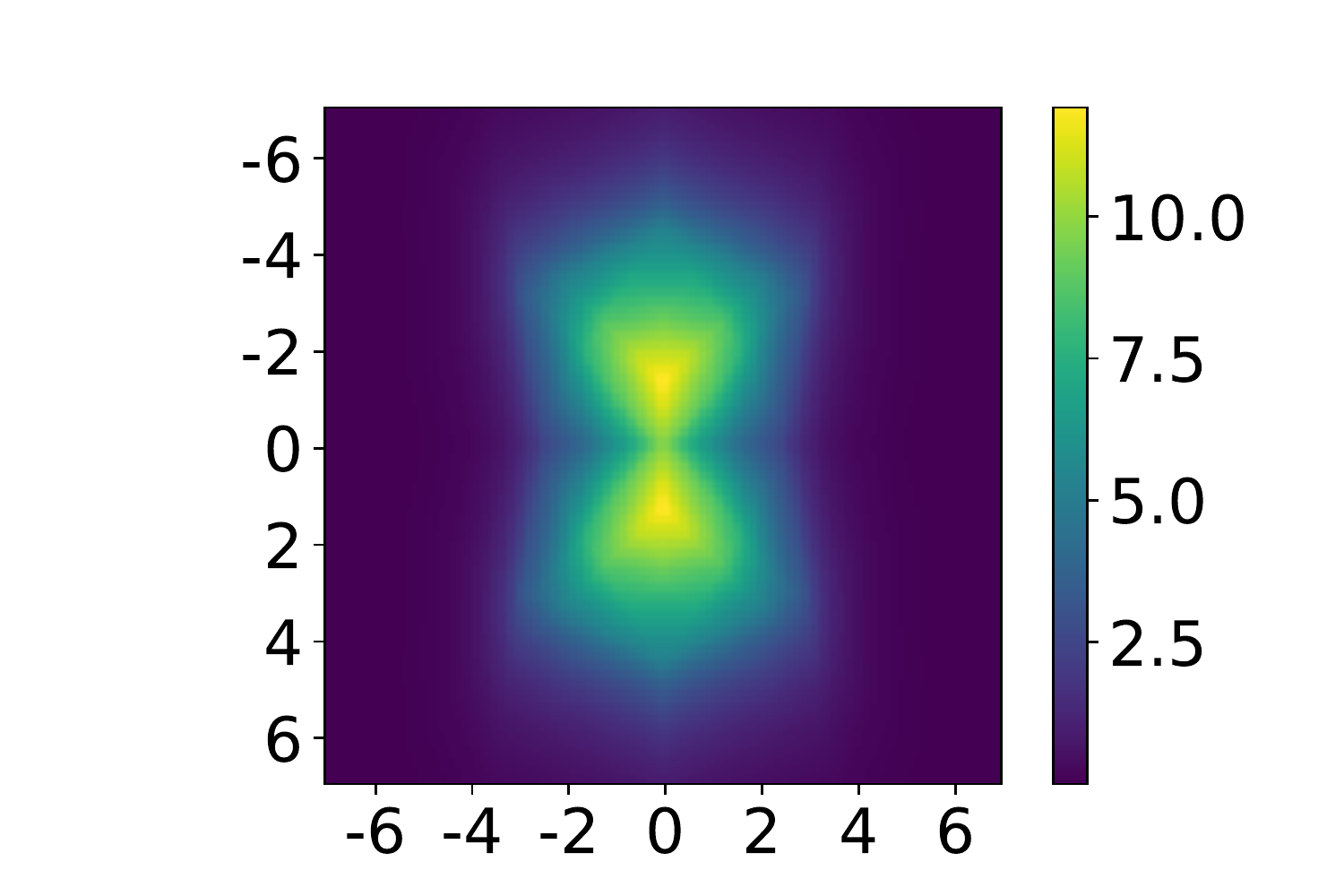} &
		\rotatebox[origin=c]{90}{{CIFAR-10}} &
		{\includegraphics[width=\figwidth, trim={2.5cm 0cm 1cm 0cm}, clip, align=c]{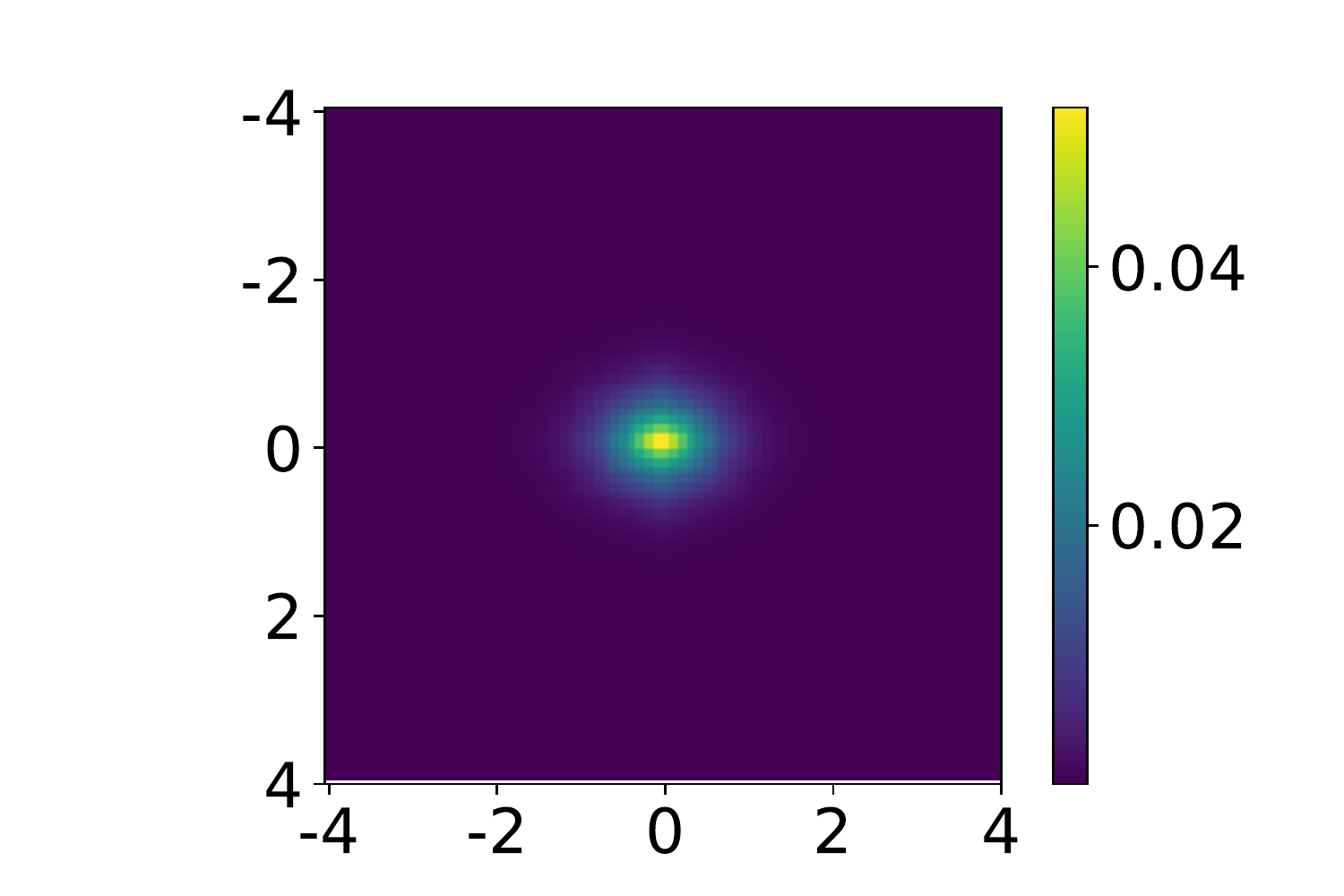}} &
		\includegraphics[width=\figwidth, trim={2.5cm 0cm 1cm 0cm}, clip, align=c]{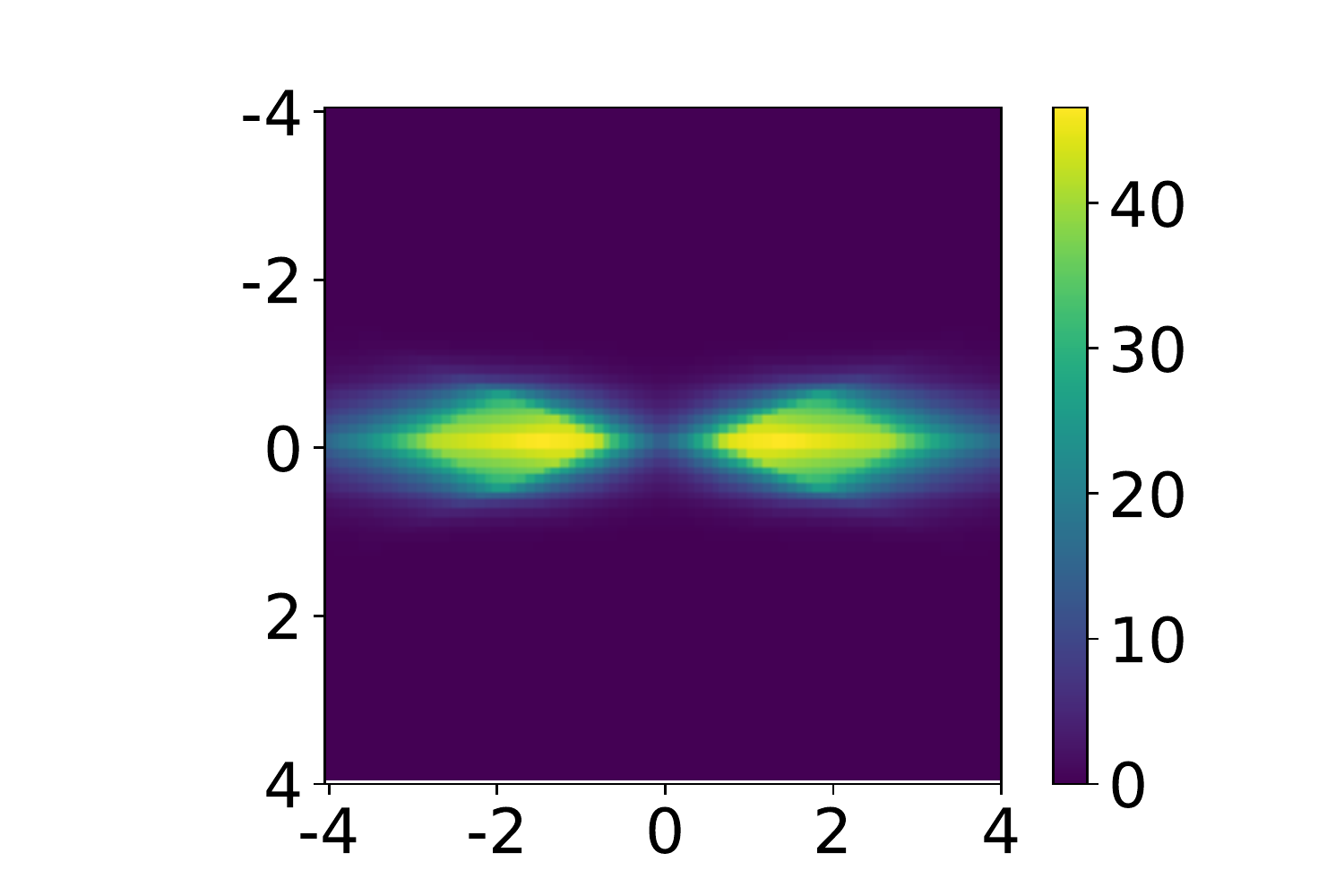} &
		\includegraphics[width=\figwidth, trim={2.5cm 0cm 1cm 0cm}, clip, align=c]{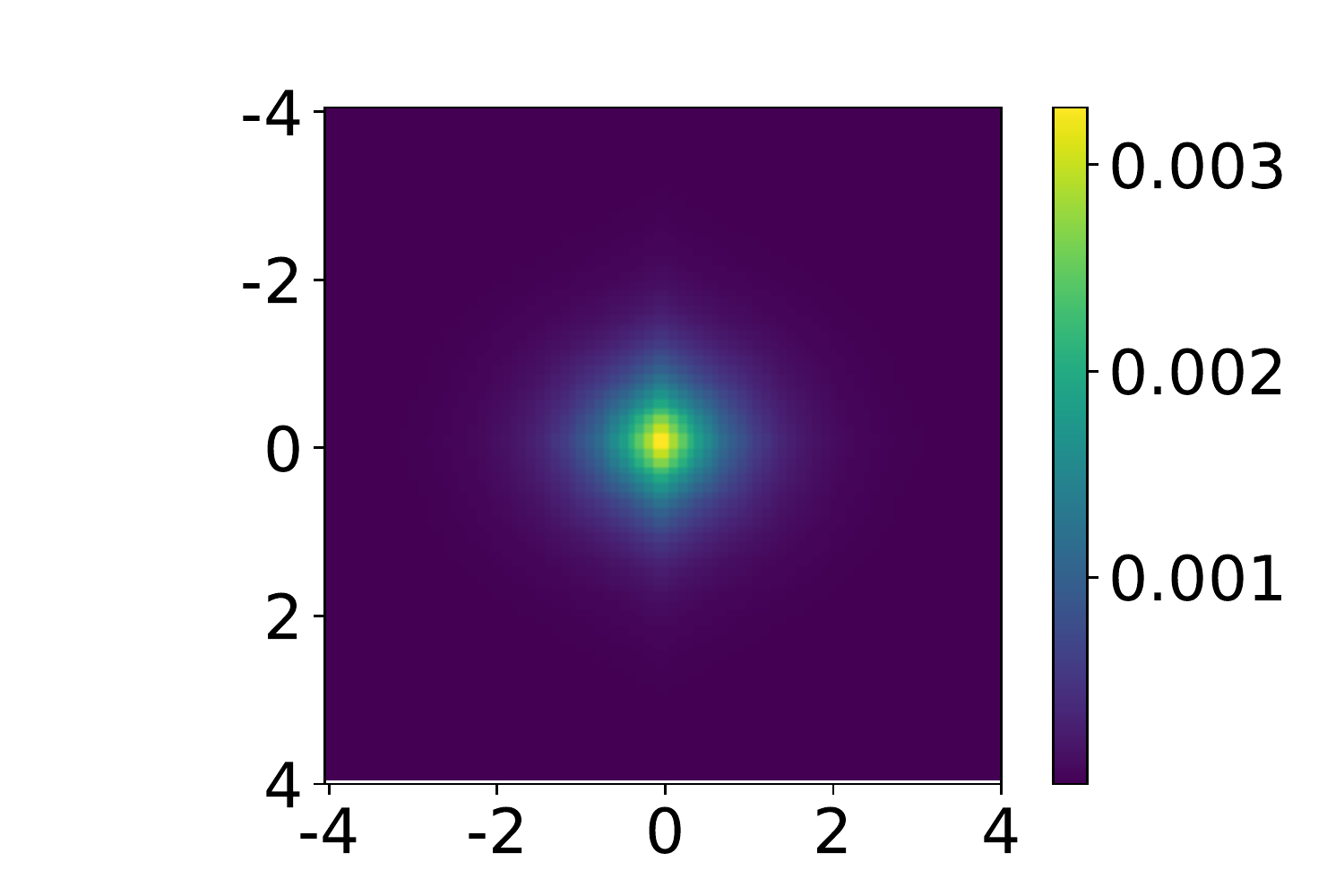} &
		\includegraphics[width=\figwidth, trim={2.5cm 0cm 1cm 0cm}, clip, align=c]{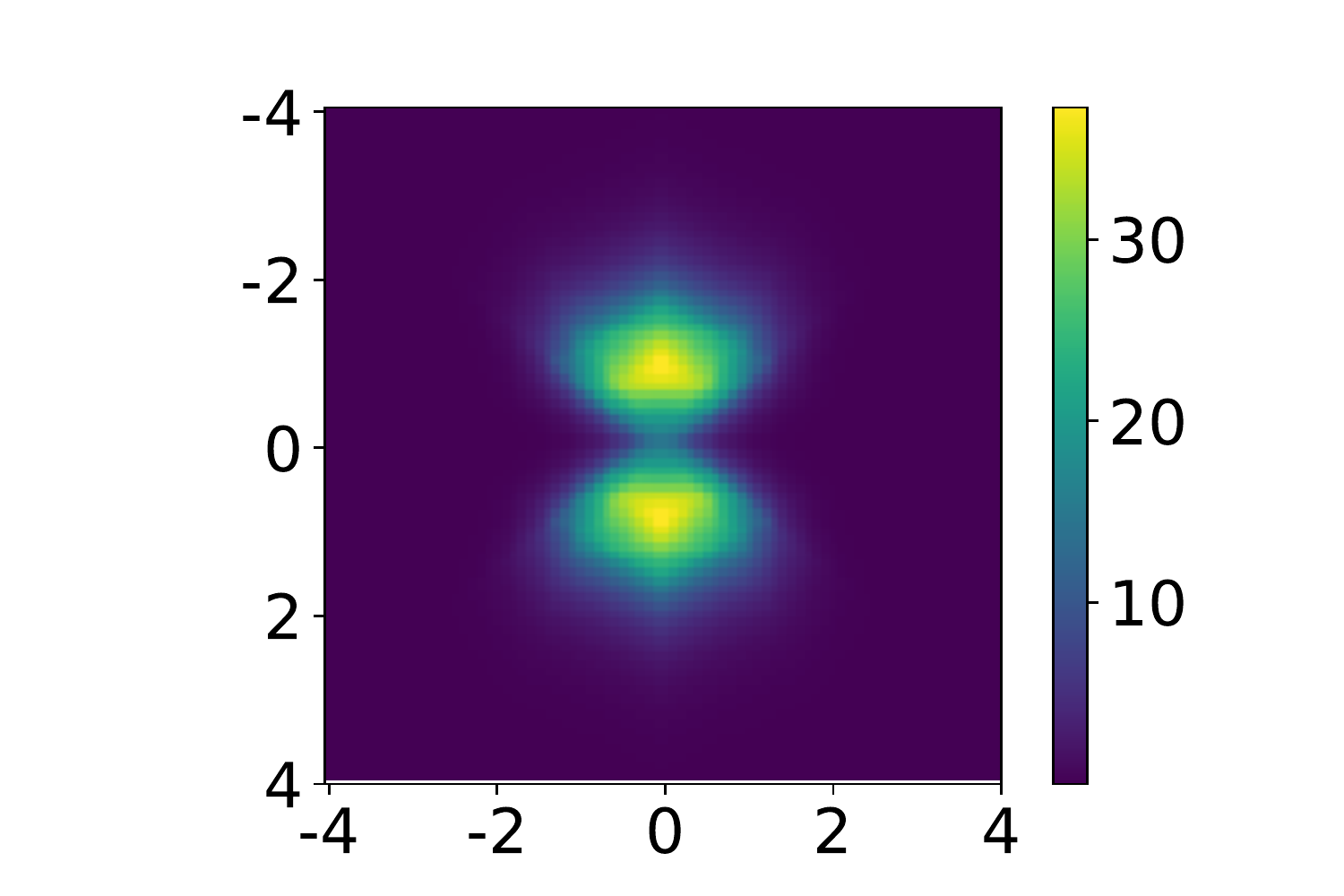} \\
		\hline
	\end{tabular}
	\vspace{1pt}
	\caption{\small Examples of predicted edges for MNIST and CIFAR-10 using the models with $L=4$ learned edges (Eq.~\ref{eq:edge_predict}). These predictions can be interpreted as filters centred at some node $v_i$, so that values along both axes denote distances from node $v_i$ to other nodes $v_j$ and the intensity denotes the strength of a connection with that node. As we can see, our models learn filters with variable intensity depending on the direction to better capture features in images. For comparison, a spatial edge (Eq.~\ref{eq:spatial_edge}) is shown on the left, which  has the same value across all directions limiting model's flexibility. }
	\vspace{-5pt}
	\label{fig:pred_edges}
\end{figure}

\densepar{Graph formation for images.}
For MNIST, we compute a hierarchy of 75, 21, and 7 SLIC~\cite{achanta2012slic} superpixels and use mean superpixel intensity and coordinates as node features, so $X \in \mathbb{R}^{N\times 3}$. For CIFAR-10, we add one more level of 150 superpixels and also use coordinate features, so $X \in \mathbb{R}^{N\times 5}$. For PASCAL, due to its more challenging images, we further add two more levels of 1,000 and 300 superpixels, so $X \in \mathbb{R}^{N\times 5}$. Note, the number of superpixels provided above are upper bounds we impose on the SLIC algorithm.
For low resolution experiments with GCNs, we build graphs from pixels and their coordinates, that is the graph structure is the same across all examples in the dataset.

\vspace{-5pt}
\subsection{Results}\label{sec:results}

\begin{figure}[t!]
	\small
	\setlength\tabcolsep{2pt}
	\begin{tabular}{ccc}
		{\includegraphics[width=0.32\textwidth, align=c, trim={0cm 0.5cm 0.5cm 0cm}, clip]{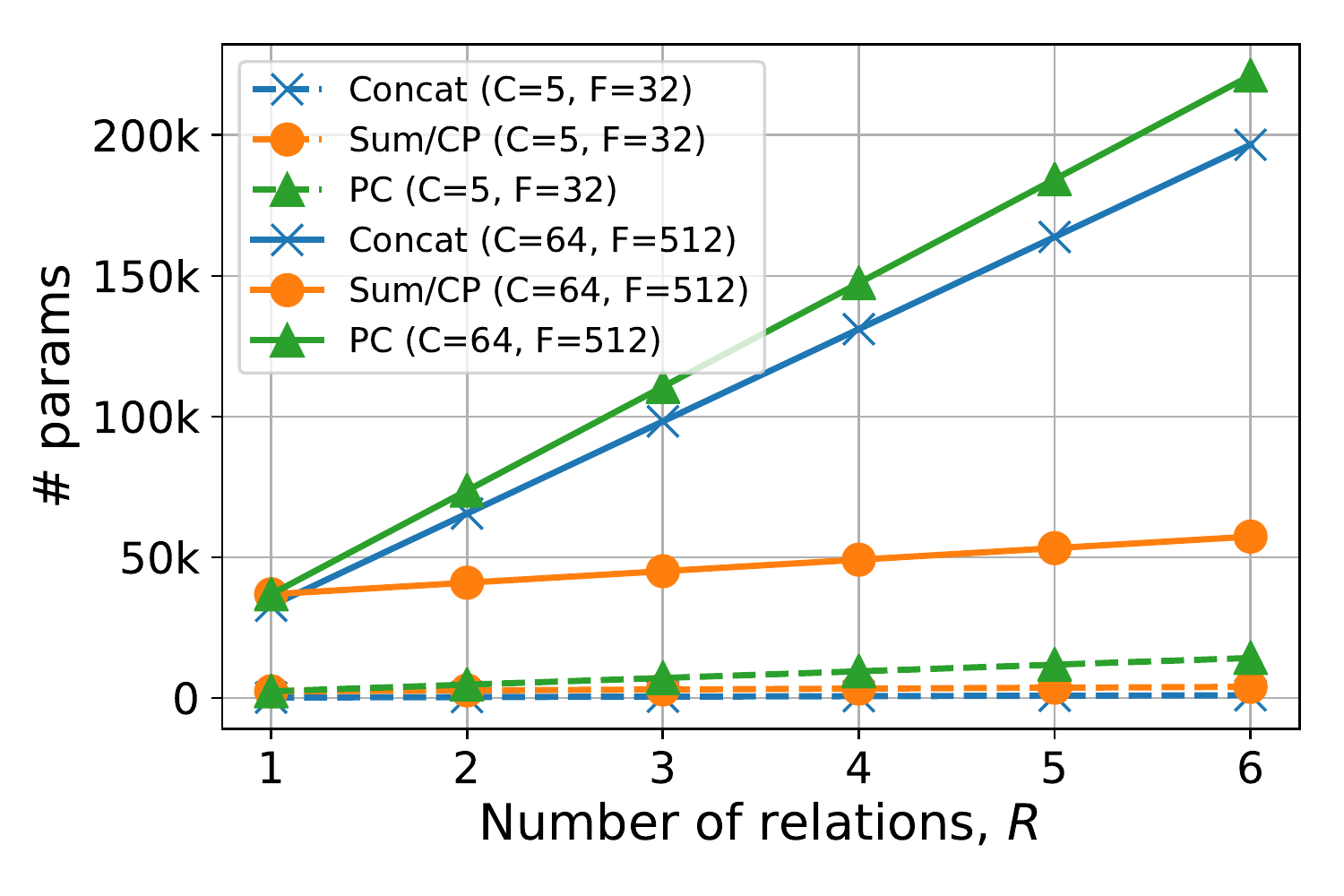}} &
		{\includegraphics[width=0.32\textwidth, align=c, trim={0cm 0cm 0.5cm 0cm}, clip]{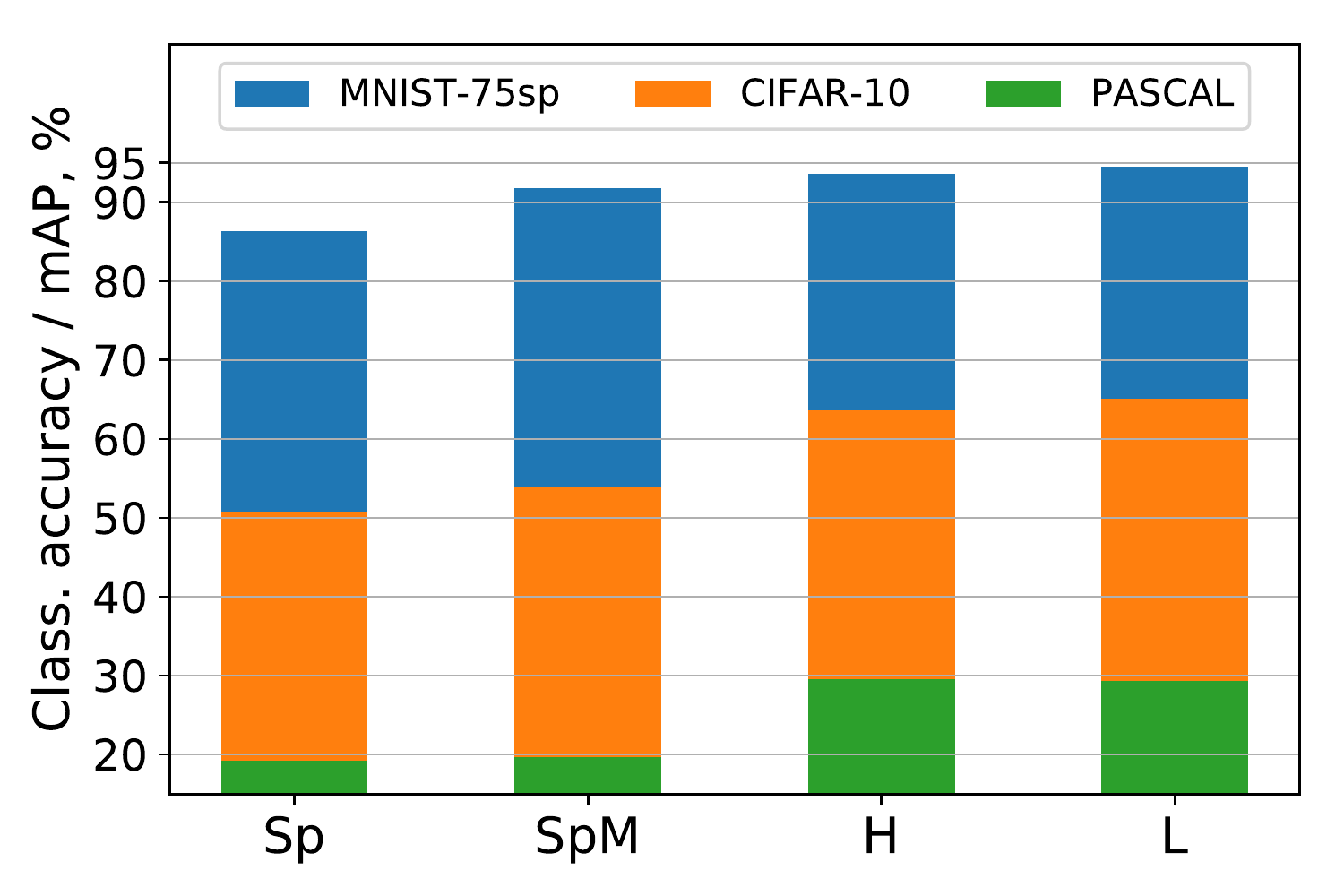}} & 
		\includegraphics[width=0.32\textwidth, trim={0cm 0cm 0cm 0cm}, clip, align=c]{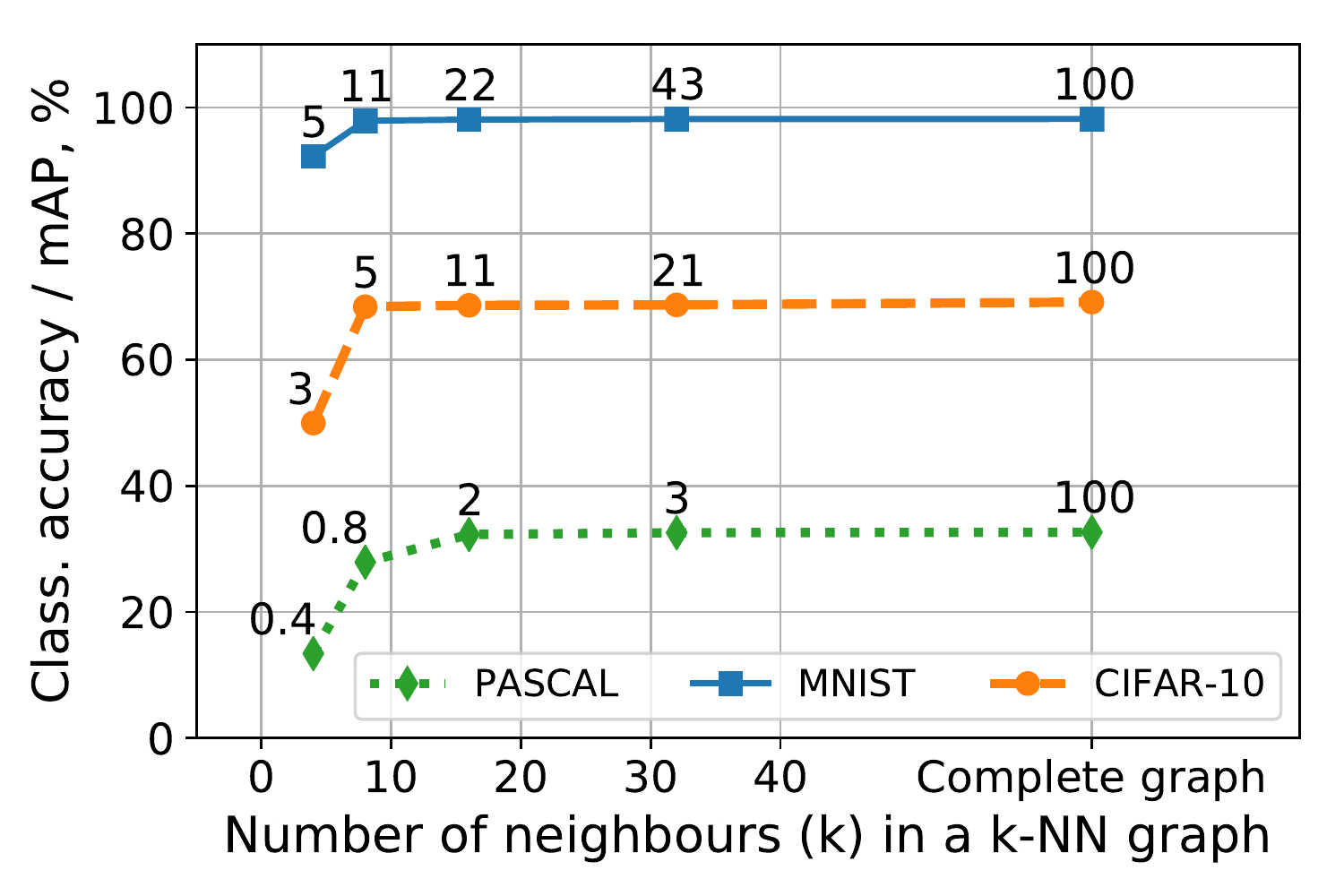} \\
		\textit{(a)} & \textit{(b)} & \textit{(c)}
	\end{tabular}
	\caption{\small \textit{(a)} Number of trainable parameters, \# params, in a graph convolutional layer as a function of the number of relations, $R$. Fusion methods based on trainable projections, including  those proposed in our work, have ``\# params'' comparable to the baseline concatenation method while being more powerful in terms of classification (see Tables~\ref{table:edge_fusions} and~\ref{table:image_class_results}). \textit{(b)} Comparison of single edge types, where learned and hierarchical edges outperform spatial edges. \textit{(c)} Effect of graph sparsification on classification accuracy. Numeric labels over markers denote sparsity of a graph (\%). Notice little or no decrease in accuracy for sparse graphs, even if they have only 2-10\% non-zero edges.}
	\label{fig:edge_fusions_sparsity}
\end{figure}

The image classification results on the MNIST, CIFAR-10 and PASCAL datasets are presented in Table~\ref{table:image_class_results}. We first observe that among GCN baselines, MoNet~\cite{monti2017geometric} and ChebyNet~\cite{defferrard2016convolutional} show comparable performance, significantly outperforming GCN~\cite{kipf2016semi}, which can be explained by very local filters in the latter.
Next, the hierarchical (H) and learned (L or L4, i.e. models with $L=4$ learnable edges) connections proposed in this work are shown to be complementary, substantially improving both GCNs and ChebyNets with a single spatial edge type. Additional qualitative and quantitative analysis is provide in Table~\ref{table:edge_fusions}, Figures~\ref{fig:pred_edges} and~\ref{fig:edge_fusions_sparsity}.
By combining hierarchical and learned edge types (H-L, H-L4) and adding multiscale filters, we achieve a further gain in accuracy while keeping the number of trainable parameters low compared to ConvNets, MoNet and ChebyNets with large $K$. Importantly, our multirelational GCNs also show better results than ConvNets with a low resolution input. 

In Table~\ref{table:image_class_results}, for ChebyNet we report results using the best $K$ chosen from the range [2, 20]: $K=10,3,2$ for MNIST, CIFAR-10 and PASCAL respectively in case of a baseline ChebyNet and $K=4,4,2$ in case of H-L-ChebyNet.  
Among evaluated fusion methods, CP and additive (S) work best for MNIST, whereas PC and CP dominate for CIFAR-10 and PASCAL.\looseness=-1

We also evaluate a baseline GCN on low resolution images to highlight the importance of SLIC superpixels compared to downsampled pixels (Table~\ref{table:image_class_results}). Surprisingly, SLIC features provide only a moderate improvement compared to pixels bringing us to two conclusions. First, average intensity values of superpixels and their coordinates are rather weak features that carry limited geometry and texture information. Stronger features can be boundaries of superpixels, but efficient and effective modeling of such information remains an open question. 
Second, we hypothesize that on full resolution images GCNs can potentially reach the performance of ConvNets or even outperform them, while having appealing properties, such as in~\cite{khasanova2017graph}. However, to scale GCNs to such large graphs, two approaches should be considered: 1) fast and effective pooling methods~\cite{ying2018hierarchical, graphunet2018}; 2) sparsification of graphs, which we evaluated and compared to complete graphs used in our experiments (Figure~\ref{fig:edge_fusions_sparsity} \textit{(c)}).

\begin{table}[t!]
	\scriptsize
	\begin{center}
		\begin{tabular}{p{0.01cm}p{0.4cm}llccc}
			& \multicolumn{2}{l}{\textbf{Fusion method}} & \textbf{\# params} & $K=1$ (GCN) & $K=2$ & $K=7$ \Bstrut \\
			\hline
			\multirow{3}{*}{\rotatebox[origin=c]{90}{$R=1$}} & (Sp) & Spatial edge (Eq.~\ref{eq:graph_conv},\ref{eq:spatial_edge}) & \multirow{3}{*}{\scriptsize{$C\times K\times F$}} & 86.36\std{0.95} & 97.08\std{0.11} & 98.17\std{0.02} \Tstrut \\
			& (SpM) & Spatial multiscale (Eq.~\ref{eq:graph_conv},\ref{eq:spatial_edge}) & & 91.74\std{0.28} & 97.38\std{0.09} & 98.10\std{0.05} \\
			& (H) & Hier. edge (Eq.~\ref{eq:graph_conv},\ref{eq:hier_edge}) & & 93.65\std{0.07} & 96.94\std{0.07} & 97.07\std{0.07} \Bstrut \\
			\hline
			\multirow{4}{*}{\rotatebox[origin=c]{90}{$R=2$}} & (C) & Concat~\cite{knyazev2018spectral} & \scriptsize{$C\times K\times R\times F$} & 97.07\std{0.06} & 97.95\std{0.07} & 98.28\std{0.11} \Tstrut \\
			& (S) & Sum~\cite{knyazev2018spectral} & \scriptsize{$F_{\text{hid}} (C\times K\times R + F)$} & \textbf{97.93}\std{0.05} & 98.30\std{0.01} & 98.44\std{0.10} \\
			& (CP) & C-proj (Eq.~\ref{eq:concat_proj_conv}) & \scriptsize{$F_{\text{hid}} (C\times K\times R + F)$} & 97.77\std{0.08} & \textbf{98.31\std{0.09}} & \textbf{98.52\std{0.09}} \\
			& (PC) & Proj-c (Eq.~\ref{eq:proj_concat_conv}) & \scriptsize{$R F_{\text{hid}} (C\times K + F)$} & 97.87\std{0.04} & 98.28\std{0.04} & {98.47\std{0.09}} \Bstrut \\
			\hline
			& \multicolumn{3}{l}{Max gain} & 4.28 & 0.93 & 0.35 \Tstrut\\
		\end{tabular}
	\end{center}
	\caption{\small Accuracy gain achieved on MNIST-75sp (MNIST with 75 superpixels) by using both spatial and hierarchical edges ($R=2$) versus a single edge ($R=1$) depending on edge fusion methods for different number of hops, $K$, of filters.}
	\label{table:edge_fusions}
\end{table}

\newcommand{\gain}[1]{{\tiny{$\color{Green} \uparrow$#1}}}
\newcommand{\loss}[1]{{\tiny{$\color{Red} \downarrow$#1}}}

\begin{table}[tbph]
	\scriptsize
	\begin{center}
		\setlength\tabcolsep{4pt}
		\begin{tabular}{p{0.3cm}llllll}
			& \textbf{Model} &  \textbf{$R$} & \multicolumn{1}{c}{\textbf{MNIST}} &  \multicolumn{1}{c}{\textbf{CIFAR-10}} & \textbf{PASCAL} & \textbf{\# params} \Bstrut\\
			\hline
			\multirow{3}{*}{\rotatebox[origin=c]{90}{\parbox{0.7cm}{\centering \tiny \vspace{3pt} Input}}}
			& ConvNets, full res, $N_{\text{pixels}}$ & & 784 & 1024 & 150 000 \Tstrut\\
			& ConvNets/GCNs, low res, $N_{\text{pixels}}$ & & 81 & 144 & 1024 \\
			& GCNs, $N_{\text{SLIC\_superpixels}}$ & & $\leq$75 & $\leq$150 & $\leq$1000 \Bstrut\\
			\hline
			\multirow{3}{*}{{\rotatebox[origin=c]{90}{{\parbox{0.7cm}{\centering \tiny Pixels: baselines}}}}} &
			ConvNet~\cite{lecun1998gradient}, full res & $-$ & 99.42\std{0.01} & 83.77\std{0.21} & 41.42\std{0.51} & 320-360k \Tstrut\\
			& ConvNet~\cite{lecun1998gradient}, low res & $-$ & 97.09\std{0.10} & 72.68\std{0.40} & 32.69\std{0.43} & 320-330k \\
			& GCN~\cite{kipf2016semi}, low res & 1 & 81.36\std{0.41} & 50.57\std{0.14} & 18.57\std{0.10} & 42k\\
			\multirow{3}{*}{\rotatebox[origin=c]{90}{{\parbox{0.7cm}{\centering \tiny SLIC: baselines}}}}
			& GCN~\cite{kipf2016semi} & 1 & 86.29\std{0.44} & 51.51\std{0.55} & 19.24\std{0.06} & 42k \Tstrut\\
			& MoNet~\cite{monti2017geometric} & 1 & 96.64\std{2.01} & 72.62\std{0.57} & $^\ddagger$ & 880k \\
			& ChebyNet~\cite{defferrard2016convolutional} & 1 & 98.24\std{0.03} & 68.92\std{0.23} & 32.38\std{0.38} & 80-350k\Bstrut\\			
			\hline
			\multirow{6}{*}{\rotatebox[origin=c]{90}{{\parbox{0.7cm}{\centering \tiny SLIC: ours}}}} & L-GCN & 2 & 97.64\std{0.12}$^S$ (\gain{1.05}) & 70.14\std{0.29}$^{PC}$ (\gain{1.00}) & 32.39\std{0.19}$^{PC}$ (\gain{0.08}) & 100k \Tstrut\\
			& H-GCN & 2 & 97.93\std{0.05}$^S$ (\gain{0.86}) & 69.05\std{0.14}$^{PC}$ (\gain{0.60}) & 32.18\std{0.29}$^{PC}$ (\gain{0.53}) & 100k \\
			& H-L-GCN & 3 & 98.35\std{0.09}$^S$ (\gain{0.63}) & 71.44\std{0.30}$^{CP}$ (\gain{1.81}) &  31.75\std{0.74}$^{PC}$ (\gain{0.54}) & 145k  \\
			& L4-GCN & 5 & 98.42\std{0.12}$^{CP}$ (\gain{0.10}) & 72.67\std{0.36}$^{S,C}$ (\tiny{0.00}) & {33.01}\std{0.49}$^{C}$ (\loss{0.38}$^{PC}$) & 185k \\
			& H-L4-GCN & 6 & 98.66\std{0.03}$^S$ (\gain{0.22}) & 72.89\std{0.70}$^{CP}$ (\gain{0.89}) & 32.61\std{0.45}$^{C}$ (\loss{2.18}$^S$)  & 220k \\
			& H-L-ChebyNet & 3 & \textbf{98.68}\std{0.05}$^{CP}$ (\gain{0.18}) & \textbf{73.18\std{0.52}}$^{PC}$ (\gain{2.40})  & \textbf{34.46\std{0.47}}$^{PC}$ (\gain{1.46}) & 200k \\
			\hline
		\end{tabular}
	\end{center}
	\caption{\small Image classification results: accuracy for MNIST and CIFAR-10 and mAP for PASCAL (\%). Superscripts denote the best fusion method among the four studied methods (See Table~\ref{table:edge_fusions} for notation). $^\ddagger$ - unable to evaluate on our infrastructure due to high computational cost. \# params is the total number of trainable parameters in a model (largest across rows). An up-arrow (\gain{}) shows the gain compared to the baseline concatenation fusion; a down-arrow (\loss{}) shows the loss of accuracy.}
	\label{table:image_class_results}
\end{table}

\subsection{Discussion}\label{sec:discuss}

Our method relies on Graph Convolutional Networks (GCN)~\cite{kipf2016semi} and its multiscale variant ChebyNet~\cite{defferrard2016convolutional}). While ChebyNet is superior to GCN in case of a single relation type due to a larger receptive field size, adding multiple relations make both methods comparable. In fact, we show that adding hierarchical edges is generally more advantageous than adding multihop ones, because hierarchy is a strong inductive bias that facilitates capturing features of spatially remote, but hierarchically close nodes (Table~\ref{table:edge_fusions}). Learned edges also improve on spatial ones, which are defined heuristically in Eq.~\ref{eq:edge_predict} and therefore might be suboptimal.

Closely related to our work, \cite{monti2017geometric} formulated the generalized graph convolution model (MoNet) based on a trainable transformation to pseudo-coordinates, giving rise to anisotropic kernels and excellent results in visual tasks. However, we found our models with multiple relations to be better (Table~\ref{table:image_class_results}). Notably, the computational cost (both memory and speed) of MoNet is higher than for any of our models due to the costly patch operator in~\cite{monti2017geometric}, so we could not perform experiments on PASCAL with 1000 superpixels due to limited GPU memory. The argument previously made in favour of MoNet against spectral methods, including ChebyNet, was the sensitivity of spectral convolution methods to changes in graph size and structure. We contradict this argument and show strong performance of ChebyNet.

Another method, SplineCNN~\cite{fey2018splinecnn}, is similar to MoNet and is also based on pseudo-coordinates, but we leave studying this method for future work.
Note that performance of MoNet and SplineCNN on general graph classification problems where coordinates are not well defined is inferior compared to ChebyNet~\cite{knyazev2018spectral}.

Finally, a family of methods based on graph kernels~\cite{shervashidze2011weisfeiler, kriege2016valid} shows strong results on some non-visual graph classification datasets, but their application is limited to small scale graph problems with discrete node features, whereas we have real-valued features. Scalable extensions of kernel methods to graphs with continuous features were proposed~\cite{niepert2016learning, yanardag2015deep}, but they still tend to be less competitive than methods based on GCN and ChebyNet~\cite{knyazev2018spectral}.

\section{Conclusion}\label{sec:conclusion}
We address several limitations of current graph convolutional networks and show improved graph classification results on a number of image datasets. First, we formulate the classification problem in terms of multigraphs, and extend edge fusion methods based on trainable projections. Second, we propose hierarchical edges and a way to learn new edges in a graph jointly with a graph classification model. Our results show that spatial, hierarchical, learned and multihop edges have a complimentary nature, improving accuracy when combined. We show that our models can outperform standard convolutional networks in experiments with low resolution images, which should pave the way for future research in that direction.

\subsubsection*{Acknowledgments}
This research was developed with funding from the Defense Advanced Research Projects Agency (DARPA). The views, opinions and/or findings expressed are those of the author and should not be interpreted as representing the official views or policies of the Department of Defense or the U.S. Government.
The authors also acknowledge support from the Canadian Institute for Advanced Research and the Canada Foundation for Innovation.

\smallskip
{\small
	\bibliography{main}
}

\end{document}